\documentclass[acmtog]{acmart}

\AtBeginDocument{%
  }

\setcopyright{acmlicensed}
\acmYear{2024} 
\setcopyright{rightsretained} 
\acmConference[SIGGRAPH Conference Papers '24]{Special Interest Group on Computer Graphics and Interactive Techniques Conference Conference Papers '24}{July 27-August 1, 2024}{Denver, CO, USA}
\acmBooktitle{Special Interest Group on Computer Graphics and Interactive Techniques Conference Conference Papers '24 (SIGGRAPH Conference Papers '24), July 27-August 1, 2024, Denver, CO, USA}
\acmDOI{10.1145/3641519.3657428}
\acmISBN{979-8-4007-0525-0/24/07}

\usepackage{colortbl}
\usepackage{graphicx}
\usepackage{subcaption}
\usepackage{xspace}
\usepackage{multirow}
\usepackage{makecell}
\usepackage{algorithm}
\newcommand{\best}{\cellcolor{tablered}}
\newcommand{\sbest}{\cellcolor{orange}}
\newcommand{\tbest}{\cellcolor{yellow}}

\newcommand{\bc}{\mathbf{c}}

\newcommand{\bh}{\mathbf{h}}\newcommand{\bH}{\mathbf{H}}

\newcommand{\bJ}{\mathbf{J}}

\newcommand{\bM}{\mathbf{M}}
\newcommand{\bn}{\mathbf{n}}\newcommand{\bN}{\mathbf{N}}

\newcommand{\bp}{\mathbf{p}}

\newcommand{\bR}{\mathbf{R}}
\newcommand{\bS}{\mathbf{S}}
\newcommand{\bt}{\mathbf{t}}
\newcommand{\bu}{\mathbf{u}}

\newcommand{\bW}{\mathbf{W}}
\newcommand{\bx}{\mathbf{x}}

\newcommand{\bSigma}{\boldsymbol{\Sigma}}

\newcommand{\cG}{\mathcal{G}}

\newcommand{\figref}[1]{Figure~\ref{#1}}
\newcommand{\secref}[1]{Section~\ref{#1}}

\newcommand{\tabref}[1]{Table~\ref{#1}}

\makeatletter
\DeclareRobustCommand\onedot{\futurelet\@let@token\@onedot}
\def\@onedot{\ifx\@let@token.\else.\null\fi\xspace}
 
\def\ie{i.e\onedot}

\def\etal{et~al\onedot}

\makeatother

\definecolor{yellow}{rgb}{1, 1, 0.7}
\definecolor{orange}{rgb}{1, 0.85, 0.7}
\definecolor{tablered}{rgb}{1, 0.7, 0.7}
\definecolor{red}{rgb}{1, 0, 0}

\definecolor{wincolor}{rgb}{0.85, 0.0, 0.0}

\definecolor{darkyellow}{rgb}{0.8, 0.8, 0.5}
\definecolor{darkred}{rgb}{0.7, 0.3, 0.3}
\definecolor{darkgreen}{rgb}{0.3, 0.7, 0.3}
\definecolor{green}{rgb}{0, 1.0, 0}
\definecolor{pink}{rgb}{1, 0.4, 0.7}

\newcommand{\boldparagraph}[1]{\paragraph{#1:}}

\begin{document}

\title{2D Gaussian Splatting for Geometrically Accurate Radiance Fields}


\author{Binbin Huang}
\email{huangbb@shanghaitech.edu.cn}
\affiliation{
\institution{ShanghaiTech University}
\city{Shanghai}
\country{China}
}

\author{Zehao Yu}
\email{zehao.yu@uni-tuebingen.de}
\affiliation{
\institution{University of Tübingen}
\institution{Tübingen AI Center}
\city{Tübingen}
\country{Germany}
}

\author{Anpei Chen}
\email{anpei.chen@uni-tuebingen.de}
\affiliation{
\institution{University of Tübingen}
\institution{Tübingen AI Center}
\city{Tübingen}
\country{Germany}
}

\author{Andreas Geiger}
\email{a.geiger@uni-tuebingen.de}
\affiliation{
\institution{University of Tübingen}
\institution{Tübingen AI Center}
\city{Tübingen}
\country{Germany}
}

\author{Shenghua Gao}
\email{gaoshh@shanghaitech.edu.cn}
\affiliation{
\institution{ShanghaiTech University}
\city{Shanghai}
\country{China}
}

\begin{abstract}
3D Gaussian Splatting (3DGS) has recently revolutionized radiance field reconstruction, achieving high quality novel view synthesis and fast rendering speed. However, 3DGS fails to accurately represent surfaces due to the multi-view inconsistent nature of 3D Gaussians. We present 2D Gaussian Splatting (2DGS), a novel approach to model and reconstruct geometrically accurate radiance fields from multi-view images. Our key idea is to collapse the 3D volume into a set of 2D oriented planar Gaussian disks. Unlike 3D Gaussians, 2D Gaussians provide view-consistent geometry while modeling surfaces intrinsically. To accurately recover thin surfaces and achieve stable optimization, we introduce a perspective-accurate 2D splatting process utilizing ray-splat intersection and rasterization. Additionally, we incorporate depth distortion and normal consistency terms to further enhance the quality of the reconstructions. We demonstrate that our differentiable renderer allows for noise-free and detailed geometry reconstruction while maintaining competitive appearance quality, fast training speed, and real-time rendering. 
\end{abstract}



\begin{teaserfigure}
\begin{flushleft}
\vspace{-0.3cm}
{\large \textcolor{magenta}{\texttt{\href{https://surfsplatting.github.io}{https://surfsplatting.github.io}}}}\\
\vspace{0.1cm}
\begin{center}
\centering
    \includegraphics[width=0.99\linewidth]{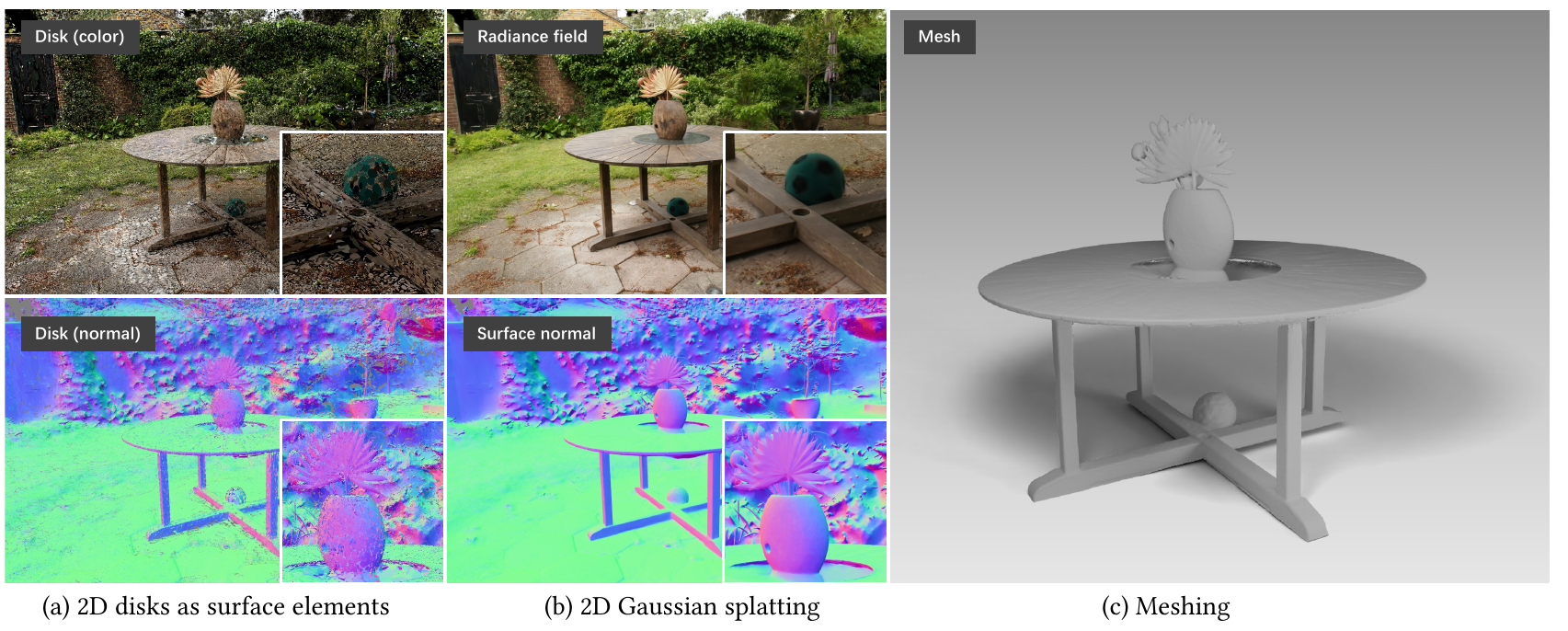}
    \caption{Our method, \textit{2DGS}, (a) optimizes a set of 2D oriented disks to represent and reconstruct a complex real-world scene from multi-view RGB images. These optimized 2D disks are tightly aligned to the surfaces. 
    (b) With 2D Gaussian splatting, we allow real-time rendering of high quality novel view images with view consistent normals and depth maps. (c) Finally, our method provides detailed and noise-free triangle mesh reconstruction from the optimized  2D disks.} 
    \label{fig:teaser}
\end{center}
\end{flushleft}
\end{teaserfigure}

\maketitle

\section{Introduction}
Photorealistic novel view synthesis (NVS) and accurate geometry reconstruction stand as pivotal long-term objectives in computer graphics and vision. Recently, 3D Gaussian Splatting (3DGS)~\cite{kerbl3Dgaussians} has emerged as an appealing alternative to implicit~\cite{mildenhall2020nerf,barron2022mip} and feature grid-based representations~\cite{Barron2023ICCV,muller2022instant} in NVS, due to its real-time photorealistic NVS results at high resolutions. Rapidly evolving, 3DGS has been quickly extended with respect to multiple domains, including anti-aliasing rendering~\cite{Yu2023MipSplatting}, material modeling~\cite{shi2023gir,jiang2023gaussianshader}, dynamic scene reconstruction~\cite{yan2023streetgaussians}, and animatable avatar creation~\cite{Zielonka2023Drivable3D,qian2023gaussianavatars}. Nevertheless, it falls short in capturing intricate geometry since the volumetric 3D Gaussian, which models the complete angular radiance, conflicts with the thin nature of surfaces. 

On the other hand, earlier works~\cite{pfister2000surfels,zwicker2001ewa, zwicker2001surface} have shown surfels (surface elements) to be an effective representation of complex geometry. Surfels approximate the object surface locally with shape and shade attributes and can be derived from known geometry. They are widely used in SLAM~\cite{whelan2016elasticfusion} and other robotics tasks~\cite{schops2019surfelmeshing} as an efficient geometry representation. Subsequent advancements~\cite{yifan2019differentiable} have incorporated surfels into a differentiable framework. However, these methods typically require ground truth (GT) geometry, depth sensor data, or operate under constrained scenarios with known lighting.

\begin{figure}[t]
    \centering
    \includegraphics[width=0.95\columnwidth]{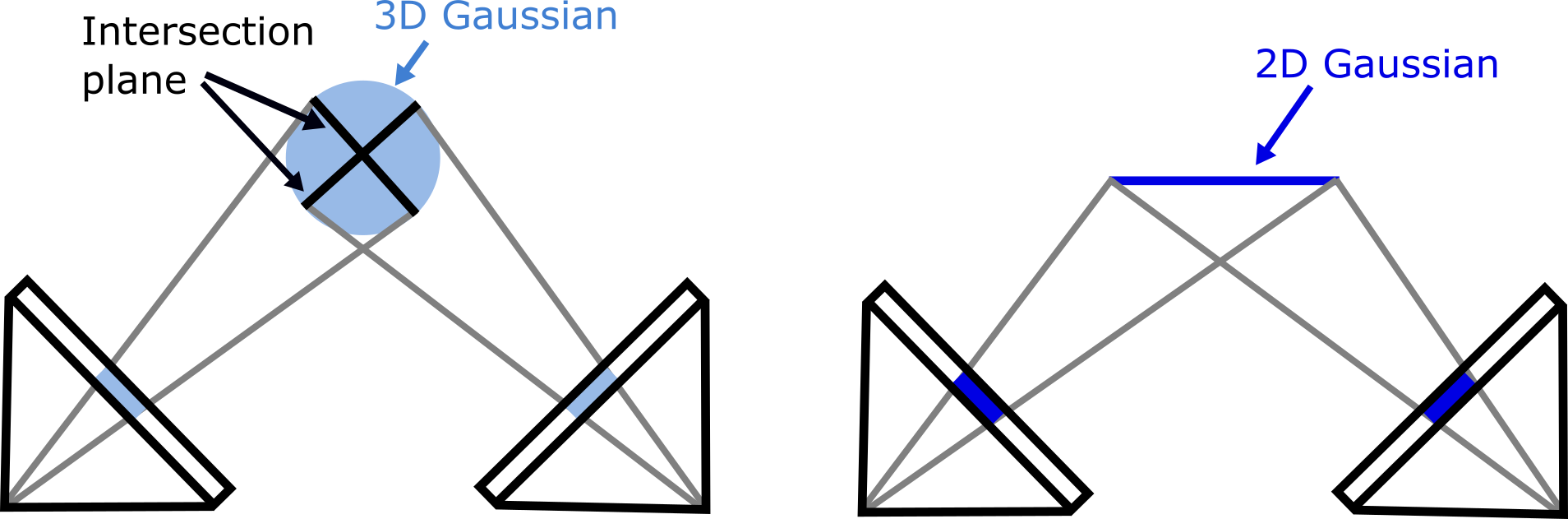}
    \caption{Comparison of 3DGS and 2DGS. 3DGS utilizes different intersection planes for value evaluation when viewing from different viewpoints, resulting in inconsistency. Our 2DGS provides multi-view consistent value evaluations.
    }
    \vspace{-10pt}
    \label{fig:2d_vs_3d}
\end{figure}

Inspired by these works, we propose 2D Gaussian Splatting for 3D scene reconstruction and novel view synthesis that combines the benefits of both worlds, while overcoming their limitations. Unlike 3DGS, our approach represents a 3D scene with 2D Gaussian primitives, each defining an oriented elliptical disk. The significant advantage of 2D Gaussian over its 3D counterpart lies in the accurate geometry representation during rendering. Specifically, 3DGS evaluates a Gaussian's value at the intersection between a pixel ray and a 3D Gaussian~\cite{keselman2022approximate,keselman2023flexible}, which leads to inconsistency depth when rendered from different viewpoints. In contrast, our method utilizes explicit ray-splat intersection, resulting in a perspective correct splatting, as illustrated in~\figref{fig:2d_vs_3d}, which in turn significantly improves reconstruction quality. Furthermore, the inherent surface normals in 2D Gaussian primitives enable direct surface regularization through normal constraints. In contrast with surfels-based models~\cite{pfister2000surfels,zwicker2001ewa,yifan2019differentiable}, our 2D Gaussians can be recovered from unknown geometry with gradient-based optimization.

While our 2D Gaussian approach excels in geometric modeling, optimizing solely with photometric losses can lead to noisy reconstructions, due to the inherently unconstrained nature of 3D reconstruction tasks, as noted in~\cite{barron2022mipnerf360,kaizhang2020,Yu2022MonoSDF}. To enhance reconstructions and achieve smoother surfaces, we introduce two regularization terms: depth distortion and normal consistency. The depth distortion term concentrates 2D primitives distributed within a tight range along the ray, addressing the rendering process's limitation where the distance between Gaussians is ignored. The normal consistency term minimizes discrepancies between the rendered normal map and the gradient of the rendered depth, ensuring alignment between the geometries defined by depth and normals. 
Employing these regularizations in combination with our 2D Gaussian model enables us to extract highly accurate surface meshes, as demonstrated in~\figref{fig:teaser}.

In summary, we make the following contributions:
\begin{itemize}
    \item We present a highly efficient differentiable 2D Gaussian renderer, enabling perspective-correct splatting by leveraging 2D surface modeling, ray-splat intersection, and volumetric integration. 
    \item We introduce two regularization losses for improved and noise-free surface reconstruction. 
    \item Our approach achieves state-of-the-art geometry reconstruction and NVS results compared to other explicit representations. 
\end{itemize}

\section{Related work}

\subsection{Novel view synthesis}
Significant advancements have been achieved in NVS, particularly since the introduction of Neural Radiance Fields (NeRF)~\cite{mildenhall2021nerf}. NeRF employs a multi-layer perceptron (MLP) to represent geometry and view-dependent appearance, optimized via volume rendering to deliver exceptional rendering quality. Post-NeRF developments have further enhanced its capabilities. For instance, Mip-NeRF~\cite{Barron2021ICCV} and subsequent works~\cite{barron2022mip,Barron2023ICCV,Hu2023ICCV} tackle NeRF's aliasing issues. Additionally, the rendering efficiency of NeRF has seen substantial improvements through techniques such as distillation~\cite{Reiser2021ICCV,yu2021plenoctrees} and baking~\cite{reiser2023merf,Hedman2021ICCV,yariv2023bakedsdf,chen2023mobilenerf}. Moreover, the training and representational power of NeRF have been enhanced using feature-grid based scene representations~\cite{Chen2022ECCV,muller2022instant,liu2020neural,Sun2022CVPR,chen2023neurbf,yu2022plenoxels}. 

Recently, 3D Gaussian Splatting (3DGS)~\cite{kerbl3Dgaussians} has emerged, demonstrating impressive real-time NVS results. This method has been quickly extended to multiple domains~\cite{Yu2023MipSplatting,yan2023streetgaussians,Zielonka2023Drivable3D,xie2023physgaussian}. In this work, we propose to ``flatten'' 3D Gaussians to 2D Gaussian primitives to better align their shape with the object surface. Combined with two novel regularization losses, our approach reconstructs surfaces more accurately than 3DGS while preserving its high-quality and real-time rendering capabilities.

\subsection{3D reconstruction}
3D Reconstruction from multi-view images has been a long-standing goal in computer vision. Multi-view stereo based methods~\cite{schoenberger2016mvs,yao2018mvsnet,Yu_2020_fastmvsnet} rely on a modular pipeline that involves feature matching, depth prediction, and fusion. In contrast, recent neural approaches~\cite{Niemeyer2020CVPR,yariv2020multiview} represent surface implicitly via an MLP~\cite{Park_2019_CVPR,Mescheder2019CVPR} , extracting surfaces post-training via the Marching Cube algorithm. Further advancements~\cite{Oechsle2021ICCV,wang2021neus,yariv2021volume} integrated implicit surfaces with volume rendering, achieving detailed surface reconstructions from RGB images. These methods have been extended to large-scale reconstructions via additional regularization~\cite{Yu2022MonoSDF,li2023neuralangelo,Yu2022SDFStudio}, and efficient reconstruction for objects~\cite{neus2}. 
Despite these impressive developments, efficient large-scale scene reconstruction remains a challenge. For instance, Neuralangelo~\cite{li2023neuralangelo} requires 128 GPU hours for reconstructing a single scene from the Tanks and Temples Dataset~\cite{Knapitsch2017}. In this work, we introduce 2D Gaussian splatting, a method that significantly accelerates the reconstruction process. It achieves similar or slightly better results compared to previous implicit neural surface representations, while being an order of magnitude faster.

\subsection{Differentiable Point-based Graphics}
Differentiable point-based rendering~\cite{insafutdinov2018unsupervised,yifan2019differentiable,aliev2020neural,wiles2020synsin,ruckert2022adop} has been explored extensively due to its efficiency and flexibility in representing intricate structures. Notably, NPBG~\cite{aliev2020neural} rasterizes point cloud features onto an image plane, subsequently utilizing a convolutional neural network for RGB image prediction. DSS~\cite{yifan2019differentiable} focuses on optimizing oriented point clouds from multi-view images under known lighting conditions. Pulsar~\cite{lassner2021pulsar} introduces a tile-based acceleration structure for more efficient rasterization. More recently, 3DGS~\cite{kerbl3Dgaussians} optimizes anisotropic 3D Gaussian primitives, demonstrating real-time photorealistic NVS results. Despite these advances, using point-based representations from unconstrained multi-view images remains challenging. In this paper, we demonstrate detailed surface reconstruction using 2D Gaussian primitives. We also highlight the critical role of additional regularization losses in optimization, showcasing their significant impact on the quality of the reconstruction.

\subsection{Concurrent works}

Since 3DGS~\cite{kerbl3Dgaussians} was introduced, it has been rapidly adapted across multiple domains. We now review the closest work in inverse rendering. These work~\cite{liang2023gs,R3DG2023,jiang2023gaussianshader,shi2023gir} extend 3DGS by modeling normals as additional attributes of 3D Gaussian primitives. 
Our approach, in contrast, inherently defines normals by representing the tangent space of the 3D surface using 2D Gaussian primitives, aligning them more closely with the underlying geometry. Additionally, the aforementioned works predominantly focus on estimating the material properties of the scene and evaluating their results for relighting tasks. Notably, none of these works specifically target surface reconstruction, the primary focus of our work. 

We also highlight the distinctions between our method and concurrent works SuGaR~\cite{guedon2023sugar} and NeuSG~\cite{chen2023neusg}. Unlike SuGaR, which approximates 2D Gaussians with 3D Gaussians, our method directly employs 2D Gaussians, simplifying the process and enhancing the resulting geometry without additional mesh refinement. NeuSG optimizes 3D Gaussian primitives and an implicit SDF network jointly and extracts the surface from the SDF network, while our approach leverages 2D Gaussian primitives for surface approximation, offering a faster and conceptually simpler solution.
\section{3D Gaussian Splatting}
\label{sec:3dgs}
Kerbl~\etal~\cite{kerbl3Dgaussians} propose to represent 3D scenes with 3D Gaussian primitives and render images using differentiable volume splatting. 
Specifically, 3DGS explicitly parameterizes Gaussian primitives via 3D covariance matrix $\bSigma$ and their location $\bp_k$: 
\begin{equation}
\cG(\bp) = \exp({-\frac{1}{2} (\bp-\bp_k)^\top \bSigma^{-1}(\bp-\bp_k)})
\label{eq:gaussian}
\end{equation}
where the covariance matrix $\bSigma = \bR\bS\bS^\top\bR^\top$ is factorized into a scaling matrix $\bS$ and a rotation matrix $\bR$.
To render an image, the 3D Gaussian is transformed into the camera coordinates with world-to-camera transform matrix $\bW$ and projected to image plane via a local affine transformation $\bJ$~\cite{zwicker2001ewa}:
\begin{equation}
\bSigma^{'} = \bJ \bW \bSigma \bW^\top \bJ^\top
\end{equation}
By skipping the third row and column of $\bSigma^{'}$, we obtain a 2D Gaussian $\cG^{2D}$ with covariance matrix $\bSigma^{2D}$. Next, 3DGS~\cite{kerbl3Dgaussians} employs volumetric alpha blending to integrate alpha-weighted appearance from front to back:
\begin{equation}
\bc(\bx) = \sum^K_{k=1} \bc_k\,\alpha_k\,\cG^{2D}_k(\bx) \prod_{j=1}^{k-1} (1 - \alpha_j\,\cG^{2D}_j(\bx))
\end{equation}
where $k$ is the index of the Gaussian primitives, $\alpha_k$ denotes the alpha values and $\bc_k$ is the view-dependent appearance. The attributes of 3D Gaussian primitives are optimized using a photometric loss.

\paragraph{Challenges in Surface Reconstruction}
Reconstructing surfaces using 3D Gaussian modeling and splatting faces several challenges. First, the volumetric radiance representation of 3D Gaussians conflicts with the thin nature of surfaces. Second, 3DGS does not natively model surface normals, essential for high-quality surface reconstruction. Third, the rasterization process in 3DGS lacks multi-view consistency, leading to varied 2D intersection planes for different viewpoints~\cite{keselman2023flexible}, as illustrated in Figure~\ref{fig:2d_vs_3d} (a). Additionally, using an affine matrix for transforming a 3D Gaussian into ray space only yields accurate projections near the center, compromising on perspective accuracy around surrounding regions~\cite{zwicker2004perspective}. Therefore, it often results in noisy reconstructions, as shown in Figure~\ref{fig:dtu_comp}.
\section{2D Gaussian Splatting}
To accurately reconstruct geometry while maintaining high-quality novel view synthesis, we present differentiable 2D Gaussian splatting (2DGS). 

\subsection{Modeling} 
\label{sec:modeling}
Unlike 3DGS~\cite{kerbl3Dgaussians}, which models the entire angular radiance in a blob, we simplify the 3-dimensional modeling by adopting ``flat'' 2D Gaussians embedded in 3D space.
{
With 2D Gaussian modeling, the primitive distributes densities within a planar disk, defining the normal as the direction of the steepest change of density.
}
This feature enables better alignment with thin surfaces. While previous methods~\cite{kopanas2021point, yifan2019differentiable} also utilize 2D Gaussians for geometry reconstruction, they require a dense point cloud or ground-truth normals as input. By contrast, we simultaneously reconstruct the appearance and geometry given only a sparse calibration point cloud and photometric supervision.

\begin{figure}[t]
    \centering
    \includegraphics[width=0.8\columnwidth]{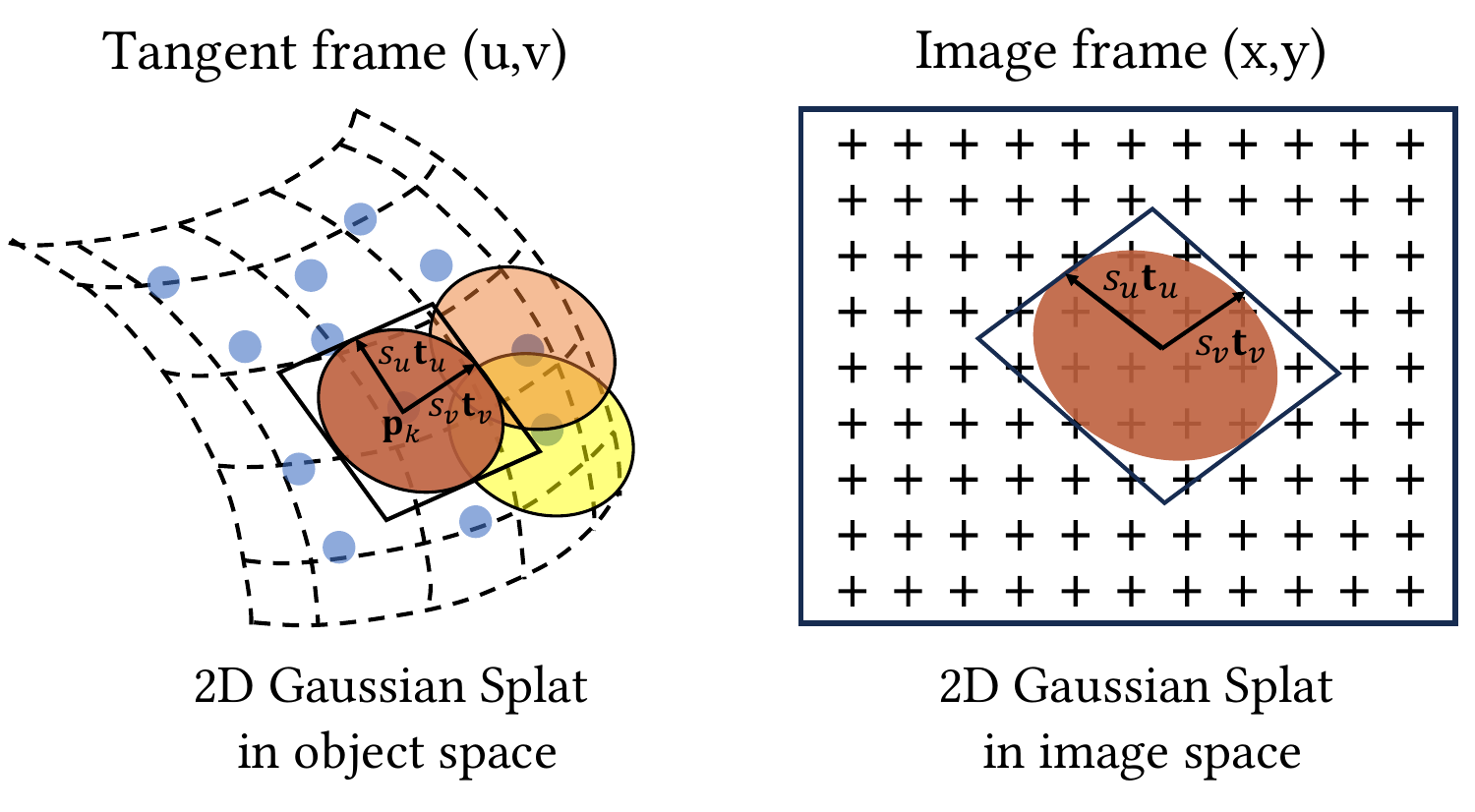}
    \caption{Illustration of 2D Gaussian Splatting. 2D Gaussian Splats are  elliptical disks characterized by a center point $\bp_k$, tangential vectors $\bt_u$ and $\bt_v$, and two scaling factors ($s_u$ and $s_v$) control the variance. Their elliptical projections are sampled through the ray-splat intersection (~\secref{sec:splatting}) and accumulated via alpha-blending in image space. 2DGS reconstructs surface attributes such as colors, depths, and normals through gradient descent. 
    }
    \label{fig:splat}
    \vspace{-18pt}
\end{figure}

As illustrated in Figure \ref{fig:splat}, our 2D splat is characterized by its central point $\bp_k$, two principal tangential vectors $\bt_u$ and $\bt_v$, and a scaling vector $\bS = (s_u,s_v)$ that controls the variances of the 2D Gaussian. Notice that the primitive normal is defined by two orthogonal tangential vectors $\bt_w = \bt_u \times \bt_v$. We can arrange the orientation into a $3 \times 3$ rotation matrix $\bR=[\bt_u,\bt_v,\bt_w]$ and the scaling factors into a $3 \times 3$ diagonal matrix $\bS$ whose last entry is zero.

A 2D Gaussian is therefore defined in a local tangent plane in world space, which is parameterized:
\begin{gather}
    P(u,v) = \bp_k + s_u \bt_u u + s_v \bt_v v = \bH(u,v,1,1)^{\mathrm{T}}\\
    \text{where} \, \bH = 
    \begin{bmatrix}
        s_u \bt_u & s_v \bt_v & \boldsymbol{0} & \bp_k \\
        0 & 0 & 0 & 1 \\
    \end{bmatrix} = \begin{bmatrix}
        \bR\bS & \bp_k \\ 
        \boldsymbol{0} & 1\\
    \end{bmatrix}
    \label{eq:plane-to-world}
\end{gather}
where $\bH \in {4\times 4}$ is a homogeneous transformation matrix representing the geometry of the 2D Gaussian. For the point $\bu=(u,v)$ in $uv$ space, its 2D Gaussian value can then be evaluated by standard Gaussian
\begin{equation}
\cG(\bu) = \exp\left(-\frac{u^2+v^2}{2}\right)
    \label{eq:gaussian-2d}
\end{equation}
The center $\bp_k$, scaling $(s_u,s_v)$, and the rotation  $(\bt_u, \bt_v)$ are learnable parameters. Following 3DGS~\cite{kerbl3Dgaussians}, each 2D Gaussian primitive has opacity $\alpha$ and view-dependent appearance $c$ parameterized with spherical harmonics.

\subsection{Splatting}
\label{sec:splatting}

{
One common strategy for rendering 2D Gaussians is to project the 2D Gaussian primitives onto the image space using the affine approximation of the perspective projection~\cite{zwicker2001surface,zwicker2001ewa}.
} However, as noted in ~\cite{zwicker2004perspective}, this projection is only accurate at the center of the Gaussian and has increasing approximation error with increased distance to the center. To address this issue, Zwicker~\etal proposed a formulation based on homogeneous coordinates. Specifically, projecting the 2D splat onto an image plane can be described by a general 2D-to-2D mapping in homogeneous coordinates. Let $\bW \in {4\times 4}$ be the combined transformation matrix from world space to screen space. The screen space points are hence obtained by
\begin{align}
    \bx = (xz, yz, z, z)^\mathrm{T} = \bW P(u,v) = \bW \bH (u,v,1,1)^\mathrm{T}
\label{eq:splat_to_screen}
\end{align}
where $\bx$ represents a homogeneous ray emitted from the camera and passing through pixel $(x, y)$ and intersecting the splat at depth $z$. 
%
{
To rasterize a 2D Gaussian, Zwicker~\etal proposed to project its conic into the screen space with an implicit method using $\mathbf{M}=(\mathbf{W}\mathbf{H})^{-1}$. 
}
However, the inverse transformation introduces numerical instability, especially when the splat degenerates into a line segment (\ie, if it is viewed from the side). To address this issue, previous surface splatting rendering methods discard such
ill-conditioned transformations using a predefined threshold~\cite{zwicker2004perspective}. However, such a scheme poses challenges within a differentiable rendering framework, as thresholding can lead to unstable optimization. To address this problem, we utilize an explicit ray-splat intersection inspired by ~\cite{sigg2006gpu}.

\paragraph{Ray-splat Intersection} We efficiently locate the ray-splat intersections by finding the intersection of three non-parallel planes, a method originally designed for specialized hardware~\cite{weyrich2007hardware}. Given an image coordinate $\bx=(x, y)$, we parameterize the ray of a pixel in the projective space as the intersection of two orthogonal planes: the x-plane and the y-plane. 
Specifically, the x-plane is defined by a normal vector $(-1, 0, 0)$ and an offset $x$. The x-plane can be represented as a 4D homogeneous plane {$\bh_x = (-1, 0, 0, x)^{\mathrm{T}}$}. Similarly, the y-plane is {$\bh_y = (0, -1, 0, y)^{\mathrm{T}}$}. Thus, the ray $\bx=(x, y)$ is determined by the intersection of the two planes. 

Next, we transform both planes into the local coordinates of the 2D Gaussian primitives, the $uv$-coordinate system. Note that transforming points on a plane using a transformation matrix $\bM$ is equivalent to transforming homogeneous plane parameters using the inverse transpose $\bM^{-\mathrm{T}}$~\cite{blinn1977homogeneous}. Therefore, applying $\bM = (\bW \bH)^{-1}$ is equivalent to $(\bW \bH)^\mathrm{T}$, eliminating explicit matrix inversion and yielding:
\begin{align}
\bh_u = (\bW \bH)^\mathrm{T} {\bh_x}  \quad \bh_v = (\bW \bH)^\mathrm{T} {\bh_y}
\label{eq:plane_transformation}
\end{align}

As introduced in ~\secref{sec:modeling}, points on the 2D Gaussian plane are represented as $(u,v,1,1)$. At the same time, the intersection point should fall in the transformed $x$-plane and $y$-plane. Thus,
\begin{align}
    \bh_u \cdot (u,v,1,1)^\mathrm{T} = \bh_v \cdot (u,v,1,1)^\mathrm{T} = 0
\end{align} 
This leads to an efficient solution for the intersection point $\bu(\bx)$:
\begin{align}
u(\bx) = \frac{\bh_u^2 \bh_v^4 - \bh_u^4 \bh_v^2}{\bh_u^1 \bh_v^2-\bh_u^2 \bh_v^1} \qquad v(\bx) = \frac{\bh_u^4 \bh_v^1 - \bh_u^1 \bh_v^4}{\bh_u^1 \bh_v^2-\bh_u^2\bh_v^1}
\label{eq:uv_intersection}
\end{align} where $\bh_u^i,\bh_v^i$ are the $i$-th parameter of the 4D plane. Note that $\bh_u^3$ and $\bh_v^3$ are always zero according to Eq.~\ref{eq:plane-to-world}. 
Once we obtain the local coordinates $(u, v)$, we can calculate the depth $z$ of the intersected points using Eq.~\ref{eq:splat_to_screen} and evaluate the Gaussian value with Eq.~\ref{eq:gaussian-2d}.

\paragraph{Degenerate Solutions} When a 2D Gaussian is observed from a slanted viewpoint, it degenerates to a line in screen space. Therefore, it might be missed during rasterization. To deal with these cases and stabilize optimization, we employ the object-space low-pass filter introduced in~\cite{botsch2005high}:
\begin{equation}
    \hat{\cG}(\bx) =\max\left\{ \cG(\bu(\bx)),  \cG(\frac{\bx-\bc}{\sigma})\right\}
\end{equation}
where $\bu(\bx)$ is given by Eq.~\ref{eq:uv_intersection} and $\bc$ is the projection of center $\bp_k$. Intuitively, $\hat{\cG}(\bx)$ is lower-bounded by a fixed screen-space Gaussian low-pass filter with center $\bc$ and radius $\sigma$. In our experiments, we set $\sigma=\sqrt{2}/2$ to ensure sufficient pixels are used during rendering.

\paragraph{Rasterization}
We follow a similar rasterization process as in 3DGS~\cite{kerbl3Dgaussians}. First, a screen space bounding box is computed for each Gaussian primitive. Then, 2D Gaussians are sorted based on the depth of their center and organized into tiles based on their bounding boxes. Finally, volumetric alpha blending is used to integrate alpha-weighted appearance from front to back:
\begin{equation}
\bc(\bx) = \sum_{i=1} \bc_i\,\alpha_i\,\hat{\cG}_i(\bu(\bx)) \prod_{j=1}^{i-1} (1 - \alpha_j\,\hat{\cG}_j(\bu(\bx)))
\end{equation}
The iterative process is terminated when the accumulated opacity reaches saturation.

\begin{figure*}[t]
\captionsetup{skip=5pt}
  \centering
  \includegraphics[width=0.99\linewidth]{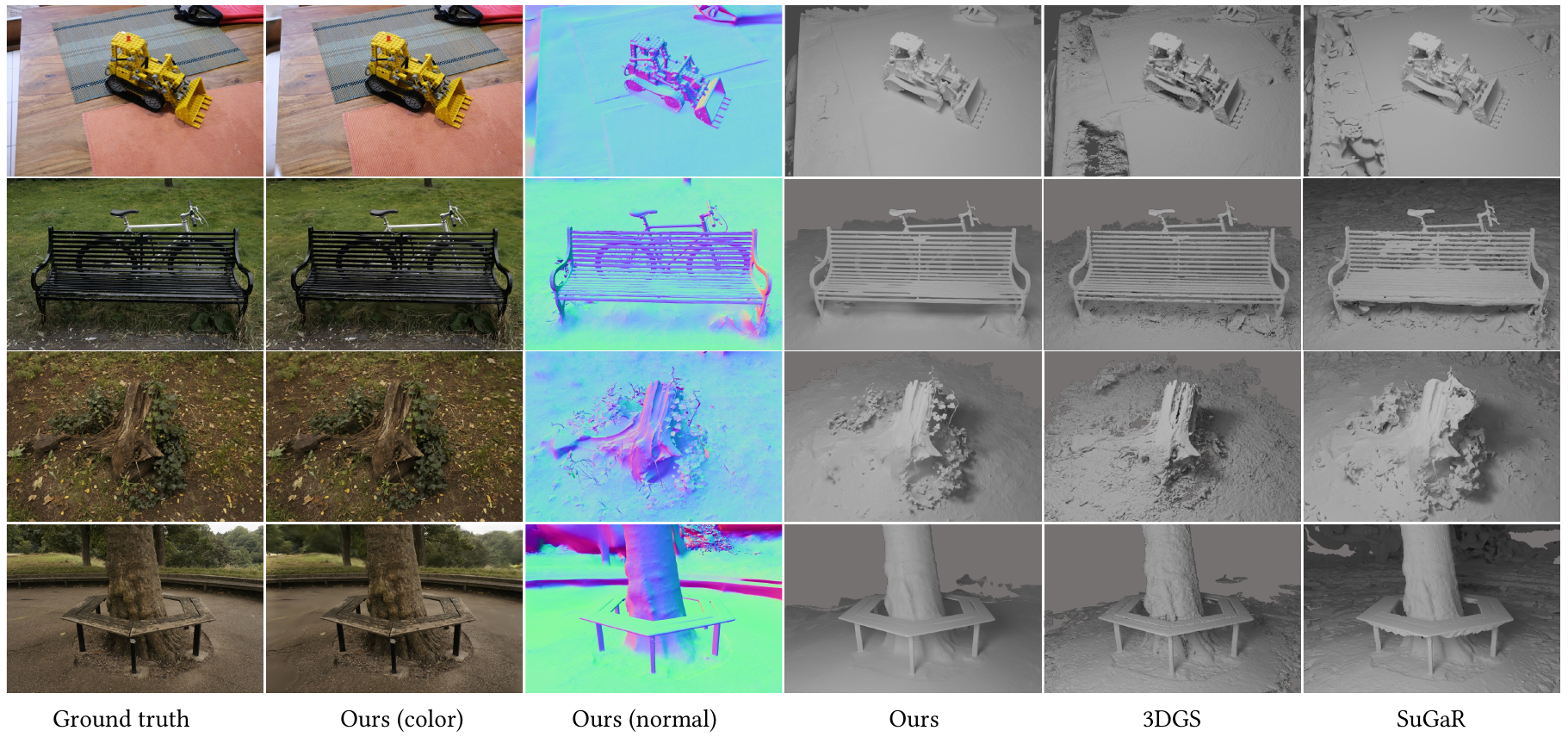}
  \caption{Visual comparisons (test-set view) between our method, 3DGS~\cite{kerbl3Dgaussians}, and SuGaR~\cite{guedon2023sugar} using scenes from an real-world dataset~\cite{barron2022mipnerf360}.
  Our method excels at synthesizing geometrically accurate radiance fields and surface reconstruction, outperforming 3DGS and SuGaR in capturing sharp edges and intricate details.
  }
  
  \label{fig:unbounded}
\end{figure*}

\section{Training}
\label{sec:optimization}
Our 2D Gaussian method, while effective in geometric modeling, can result in noisy reconstructions when optimized only with photometric losses, a challenge inherent to 3D reconstruction tasks~\cite{barron2022mipnerf360,kaizhang2020,Yu2022MonoSDF}. To mitigate this issue and improve the geometry reconstruction, we introduce two regularization terms: depth distortion and normal consistency. 

\paragraph{Depth Distortion}
Different from NeRF, 3DGS's volume rendering doesn't consider the distance between intersected Gaussian primitives. Therefore, spreading out Gaussians might result in a similar color and depth rendering. This is different from surface rendering, where rays intersect the first visible surface exactly once. To mitigate this issue, we take inspiration from Mip-NeRF360~\cite{barron2022mip} and propose a depth distortion loss to concentrate the weight distribution along the rays by minimizing the distance between the ray-splat intersections:

\begin{equation}
\mathcal{L}_{d} = \sum_{i,j}\omega_i\omega_j|z_i-z_j|
\end{equation}
where $\omega_i = \alpha_i\,\hat{\cG}_i(\bu(\bx))\prod_{j=1}^{i-1} (1 - \alpha_j\,\hat{\cG}_j(\bu(\bx)))$ is the blending weight of the $i-$th intersection and $z_i$ is the depth of the intersection points. Unlike the distortion loss in Mip-NeRF360, where $z_i$ is the distance between sampled points and is not optimized, our approach directly encourages the concentration of the splats by adjusting the intersection depth $z_i$. Note that we implement this regularization term efficiently with CUDA in a manner similar to~\cite{SunSC22_2}.

\setlength\tabcolsep{0.5em}
\begin{table*}[t]
\vspace{-2pt}
\centering
\caption{Quantitative comparison on the DTU Dataset~\cite{jensen2014large}. Our 2DGS achieves the highest reconstruction accuracy among other methods and provides $100\times$ speed up compared to the SDF based baselines.}
\resizebox{.98\textwidth}{!}{
\begin{tabular}{@{}llcccccccccccccccclcc}
\hline
 \multicolumn{3}{c}{} & 24 & 37 & 40 & 55 & 63 & 65 & 69 & 83 & 97 & 105 & 106 & 110 & 114 & 118 & 122 & & Mean & Time \\ \cline{4-18} \cline{20-21}
\multirow{3}{*}{\rotatebox[origin=c]{90}{implicit}} & NeRF~\cite{mildenhall2021nerf} & & 1.90 & 1.60 & 1.85 & 0.58 & 2.28 & 1.27 & 1.47 & 1.67 & 2.05 & 1.07 & 0.88 & 2.53 & 1.06 & 1.15 & 0.96 & & 1.49 & > 12h \\
 & VolSDF~\cite{yariv2021volume} & & \tbest 1.14 & \tbest 1.26 & \tbest 0.81 & 0.49 & 1.25 & \sbest 0.70 & \sbest 0.72 & \best 1.29 & \sbest1.18 & \best 0.70 & \sbest0.66 & \best1.08 & \tbest 0.42 & \sbest 0.61 & \tbest 0.55 & & 0.86 & >12h \\
 & NeuS~\cite{wang2021neus} & & \sbest 1.00 & 1.37 & 0.93 & \tbest 0.43 & \tbest 1.10 & \best 0.65 &  \best 0.57 & \tbest 1.48 & \best 1.09 & \tbest 0.83 & \best 0.52 & \sbest 1.20 & \best0.35 & \best 0.49 & \sbest 0.54 & & \tbest 0.84 & >12h \\
 \cline{2-2} \cline{4-18} \cline{20-21}
\multirow{3}{*}{\rotatebox[origin=c]{90}{explicit}} 
&  3DGS~\cite{kerbl3Dgaussians} & & 2.14 & 1.53 & 2.08 & 1.68 & 3.49 & 2.21 & 1.43 & 2.07 & 2.22 & 1.75 &  1.79 & 2.55 & 1.53 & 1.52 & 1.50 & & 1.96 & \tbest{11.2~m} \\
 &  SuGaR~\cite{guedon2023sugar} & & 1.47 & 1.33 & 1.13 & 0.61 & 2.25 & 1.71 & 1.15 & 1.63 & 1.62 & 1.07 & 0.79 & 2.45 & 0.98 & 0.88 & 0.79 & & 1.33 & $\sim$~1h \\
 & 2DGS-15k (Ours) && \best 0.48 & \sbest 0.92 & \sbest 0.42 & \sbest 0.40 & \sbest 1.04 & \tbest 0.83 & 0.83 & \sbest 1.36 & \tbest 1.27 & \sbest 0.76 & 0.72 & 1.63 & \sbest 0.40 & \tbest 0.76 & 0.60 &&\sbest 0.83 & \best  5.5~m \\
 & 2DGS-30k (Ours) && \best 0.48 & \best 0.91 & \best 0.39 & \best 0.39 & \best 1.01 & \tbest 0.83 & \tbest 0.81 & \sbest 1.36 & \tbest 1.27 & \sbest 0.76  & \tbest 0.70 & \tbest 1.40 &  \sbest 0.40 &  \tbest 0.76 & \best 0.52 && \best 0.80 & \tbest \sbest 10.9~m \\
 \hline
\end{tabular}
}
\vspace{-2pt}
\label{tab:dtu_result}
\end{table*}
\paragraph{Normal Consistency} 
As our representation is based on 2D Gaussian surface elements, we must ensure that all 2D splats are locally aligned with the actual surfaces. In the context of volume rendering where multiple semi-transparent surfels may exist along the ray, we consider the actual surface at the median point of intersection $\bp_s$, where the accumulated opacity reaches 0.5. We then align the splats' normal with the gradients of the depth maps as follows:
\begin{equation}
\mathcal{L}_{n} = \sum_{i} \omega_i (1-\bn_i^\mathrm{T}\bN)
\end{equation}
where $i$ indexes over intersected splats along the ray, $\omega$ denotes the blending weight of the intersection point, $\bn_i$ represents the normal of the splat that is oriented towards the camera, and $\bN$ is the normal estimated by the gradient of the depth map. Specifically, $\bN$ is computed with finite differences from nearby depth points as follows:
\begin{equation}
\mathbf{N}(x,y) = \frac{\nabla_x \bp_s \times \nabla_y \bp_s}{|\nabla_x \bp_s \times \nabla_y \bp_s|}
\label{eq:normal_depth}
\end{equation} 
By aligning the splat normal with the estimated surface normal, we ensure that 2D splats locally approximate the actual object surface. 

\paragraph{Final Loss} Finally, we optimize our model from an initial sparse point cloud using a set of posed images. We minimize the following loss function:
\begin{equation}
\mathcal{L} = \mathcal{L}_{c} + \alpha \mathcal{L}_{d}  + \beta  \mathcal{L}_{n}
\end{equation}
where $\mathcal{L}c$ is an RGB reconstruction loss combining $\mathcal{L}_1$ with the D-SSIM term from \cite{kerbl3Dgaussians}, while $\mathcal{L}_{d}$ and $\mathcal{L}_{n}$ are regularization terms. We set $\alpha=1000$ for bounded scenes, $\alpha=100$ for unbounded scenes, and $\beta=0.05$ for all scenes. 
\section{Experiments}
We now present evaluations of our 2D Gaussian Splatting reconstruction method, including appearance and geometry comparison with previous state-of-the-art implicit and explicit approaches. We then analyze the contribution of the proposed components.

\subsection{Implementation} We implement our 2D Gaussian Splatting with custom CUDA kernels, building upon the framework of 3DGS~\cite{kerbl3Dgaussians}. We extend the renderer to output depth distortion maps, depth maps and normal maps for regularizations {(See detailed computations in Appendices A and B of the supplemental material)}. During training, we increase the number of 2D Gaussian primitives following the adaptive control strategy in 3DGS. Since our method does not directly rely on the gradient of the projected 2D center, we hence project the gradient of 3D center $\bp_k$ onto the screen space as an approximation. Similarly, we employ a gradient threshold of $0.0002$ and remove splats with opacity lower than $0.05$  every $3000$ step. We conduct all the experiments on a single GTX RTX3090 GPU.

\paragraph{Mesh Extraction} To extract meshes from reconstructed 2D splats, we render depth maps of the training views using the depth value of the splats projected to the pixels and utilize truncated signed distance fusion (TSDF) to fuse the reconstruction depth maps, using Open3D~\cite{Zhou2018}. We set the voxel size to $0.004$ and the truncated threshold to 0.02 during TSDF fusion. We also extend the original 3DGS to render depth and employ the same technique for surface reconstruction for a fair comparison.

\setlength\tabcolsep{0.5em}
\begin{table}[t]
\centering
\caption{Quantitative results on the Tanks and Temples Dataset~\cite{Knapitsch2017}. We report the F1 score and training time.}
\resizebox{0.98\columnwidth}{!}{
\begin{tabular}{@{}l|ccc|ccc}
 & NeuS & Geo-Neus & Neurlangelo & SuGaR & 3DGS & Ours\\ 
 \hline
Barn & 0.29 &  0.33 &  0.70  & 0.14 & 0.13 & 0.41 \\
Caterpillar &  0.29 & 0.26 &  0.36 & 0.16 & 0.08 & 0.23\\
Courthouse &  0.17 & 0.12 &  0.28 & 0.08 & 0.09 & 0.16 \\
Ignatius &   0.83 & 0.72 &  0.89 & 0.33 & 0.04 & 0.51 \\
Meetingroom &   0.24 & 0.20 &  0.32 &  0.15 & 0.01 & 0.17\\
Truck &  0.45 &  0.45 &  0.48 &  0.26 & 0.19 & 0.45 \\ 
\hline
Mean &  0.38 & 0.35 &  0.50 & 0.19 & 0.09 & 0.32\\
Time & >24h & >24h & >24h & >1h & 14.3~m & 15.5 ~m \\
\end{tabular}
}
\label{tab:tnt}
\vspace{-10pt}
\end{table}

\begin{table}[t]
\centering
\caption{Performance comparison between 2DGS (ours), 3DGS and SuGaR on the DTU dataset~\cite{jensen2014large}. We report the averaged chamfer distance, PSNR (training-set view), reconstruction time, and model size.}
\resizebox{0.98\columnwidth}{!}{
\begin{tabular}{@{}l|cccc}
 & CD~$\downarrow$ & PSNR~$\uparrow$ & Time~$\downarrow$ & MB~(Storage) $\downarrow$\\
\hline
3DGS~\cite{kerbl3Dgaussians} & 1.96 & \textbf{35.76} & 11.2~m & 113\\
SuGaR~\cite{guedon2023sugar} & 1.33 & 34.57 & $\sim$1~h &1247 \\
2DGS-15k (Ours) & 0.83 & 33.42& \textbf{5.5~m} & \textbf{52}\\
2DGS-30k (Ours) & \textbf{0.80} & 34.52 & 10.9~m & \textbf{52}\\
\end{tabular}%
}
\vspace{-5pt}
\label{tab:dtu_perf}
\end{table}

\begin{figure}[tbh]
    \centering
    \includegraphics[width=\columnwidth]{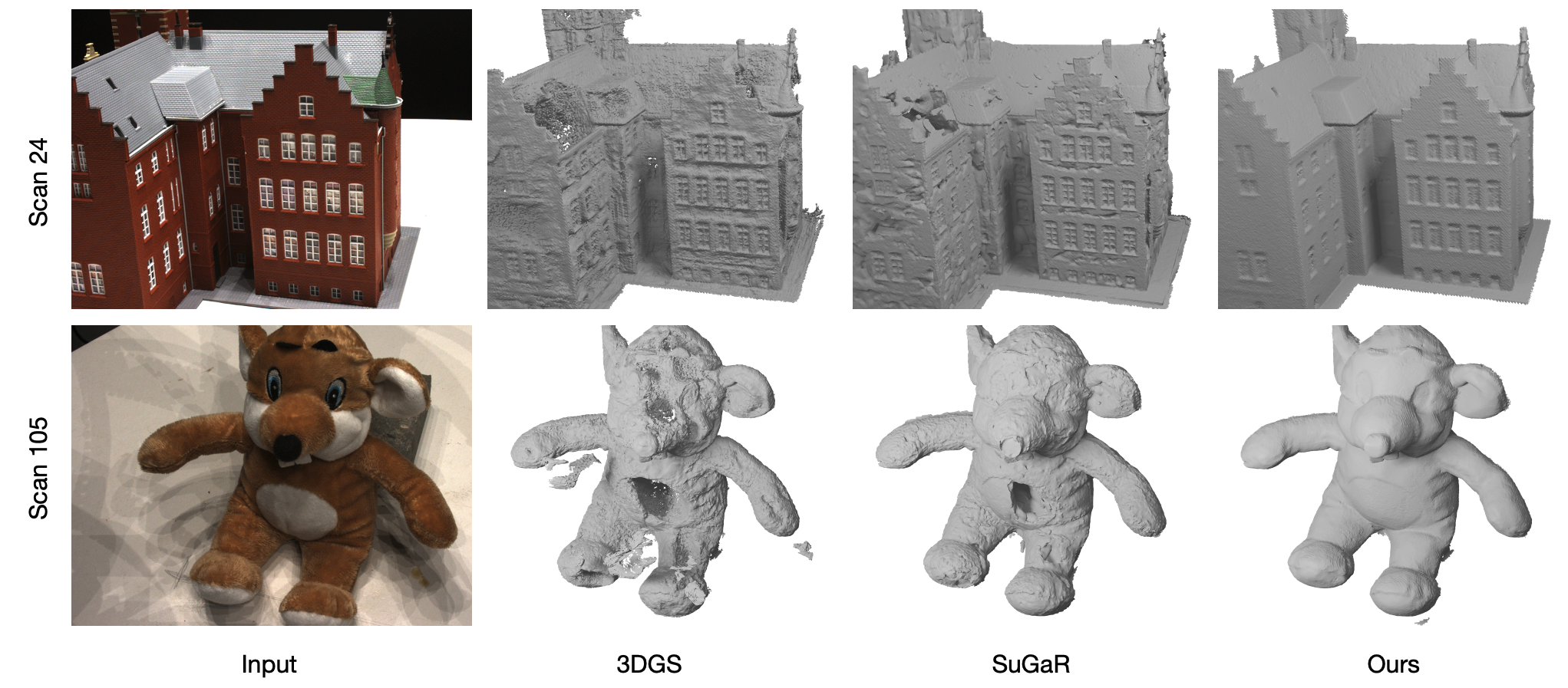}
    \caption{Qualitative comparison on the DTU benchmark~\cite{jensen2014large}. Our 2DGS produces detailed and noise-free surfaces.}
    \label{fig:dtu_comp}
\end{figure}

\begin{table}[t]
\vspace{-2pt}
\centering
\caption{Quantitative results on Mip-NeRF 360~\cite{barron2022mip} dataset. All scores of the baseline methods are directly taken from their papers whenever available. We report the performance of 3DGS, SuGaR and ours using $30k$ iterations.
}
\resizebox{0.98\columnwidth}{!}{
\begin{tabular}{@{}l|ccc|ccc}
 & \multicolumn{3}{c@{}|}{Outdoor Scene} & \multicolumn{3}{c@{}}{Indoor scene} \\ 
& PSNR~$\uparrow$ & SSIM~$\uparrow$ & LIPPS~$\downarrow$ & PSNR~$\uparrow$ & 
SSIM~$\uparrow$ & LIPPS~$\downarrow$ \\
\hline
NeRF & 21.46 & 0.458 & 0.515 & 26.84 &  0.790 & 0.370 \\
Deep Blending & 21.54 &0.524 & 0.364 & 26.40 & 0.844 & 0.261 \\
Instant NGP & 22.90 & 0.566 & 0.371 & 29.15 & 0.880 & 0.216 \\
MERF & 23.19 & 0.616 &  0.343 & 27.80 & 0.855 & 0.271 \\
BakedSDF & 22.47 & 0.585 &  0.349 & 27.06 & 0.836 & 0.258 \\
MipNeRF360 &  \sbest 24.47 & \tbest 0.691 & \tbest 0.283 &  \best 31.72 & \sbest 0.917 & \best 0.180 \\
\hline
\hline
Mobile-NeRF & 21.95 & 0.470 & 0.470 & - & - & - \\
SuGaR &  22.93 & 0.629 & 0.356 & 29.43 & 0.906 & 0.225 \\
3DGS &  \best 24.64 &  \best 0.731 &  \best 0.234 &   \sbest 30.41 &  \best 0.920 &  \sbest 0.189 \\
2DGS~(Ours) & \tbest 24.34 &  \sbest 0.717 & \sbest 0.246  & \tbest 30.40 & \tbest 0.916 & \tbest 0.195  \\
\end{tabular}
}
\label{tab:mipnerf360}
\vspace{-2pt}
\end{table}

\subsection{Comparison} 

\paragraph{Dataset}
We evaluate the performance of our method on various datasets, including
DTU~\cite{jensen2014large}, Tanks and Temples~\cite{Knapitsch2017}, and Mip-NeRF360~\cite{barron2022mip}. The DTU dataset comprises 15 scenes, each with 49 or 69 images of resolution $1600\times1200$. We use Colmap~\cite{schoenberger2016sfm} to generate a sparse point cloud for each scene and down-sample the images into a resolution of $800 \times 600$ for efficiency. We use the same training process for 3DGS~\cite{kerbl3Dgaussians} and SuGaR~\cite{guedon2023sugar} for a comparison.

\paragraph{Geometry Reconstruction}
In ~\tabref{tab:dtu_result} and ~\tabref{tab:dtu_perf}, we compare our geometry reconstruction to SOTA implicit (i.e., NeRF~\cite{mildenhall2020nerf}, VolSDF~\cite{yariv2021volume}, and NeuS~\cite{wang2021neus}), explicit (i.e., 3DGS~\cite{kerbl3Dgaussians} and concurrent work SuGaR~\cite{guedon2023sugar}) methods on Chamfer distance and training time using the DTU dataset. Our method outperforms all compared methods in terms of Chamfer distance. Moreover, as shown in~\tabref{tab:tnt}, 2DGS achieves competitive results with SDF models (i.e., NeuS~\cite{wang2021neus} and Geo-Neus~\cite{Fu2022GeoNeus}) on the TnT dataset, and significantly better reconstruction than explicit reconstruction methods (i.e., 3DGS and SuGaR). Notably, our model demonstrates exceptional efficiency, offering a reconstruction speed that is approximately 100 times faster compared to implicit reconstruction methods and more than 3 times faster than the concurrent work SuGaR. Our approach can also achieve qualitatively better reconstructions with more appearance and geometry details and fewer outliers, as shown in ~\figref{fig:dtu_comp}. Moreover, SDF-based reconstruction methods require predefining the spherical size for initialization, which plays a critical role in the success of SDF reconstruction. By contrast, our method leverages radiance field based geometry modeling and is less sensitive to initialization. {
We include the full geometry reconstruction results for both DTU and TnT in Figure~\ref{fig:supp-DTU} and Figure~\ref{fig:supp-TnT}.
}

\paragraph{Appearance Reconstruction}
Our method represents 3D scenes as radiance fields, providing high-quality novel view synthesis. In this section, we compare our novel view renderings using the Mip-NeRF360 dataset against baseline approaches, as shown in ~\tabref{tab:mipnerf360} and ~\figref{fig:unbounded}. Note that, since the ground truth geometry is not available in the Mip-NeRF360 dataset and we hence focus on quantitative comparison.  Remarkably, our method consistently achieves competitive NVS results across state-of-the-art techniques while providing geometrically accurate surface reconstruction. 
{We include the appearance rendering results in Figure~\ref{fig:supp-rendering}.}

\begin{table}[t]
\caption{Quantitative studies for the regularization terms and mesh extraction methods on the DTU dataset. }
\centering
\vspace{-2pt}
\resizebox{.96\linewidth}{!}{
\begin{tabular}{@{}l|ccc}
 & Accuracy~$\downarrow$ & Completion ~$\downarrow$ & Average~$\downarrow$ \\
\hline
A. w/o normal consistency & 1.35 & 1.13 & 1.24\\
B. w/o depth distortion & 0.89 & 0.87 & 0.88 \\
\hline
C. w / expected depth & 0.88 & 1.01 & 0.94 \\
D. w / SPSR & 1.25 & 0.89 & 1.07 \\
\hline
E. Full Model & 0.79 & 0.86 & 0.83\\
\end{tabular}%
}
\label{tab:ablation}
\vspace{-5pt}
\end{table}

\begin{figure}[t]
    \centering
    \includegraphics[width=.96\columnwidth]{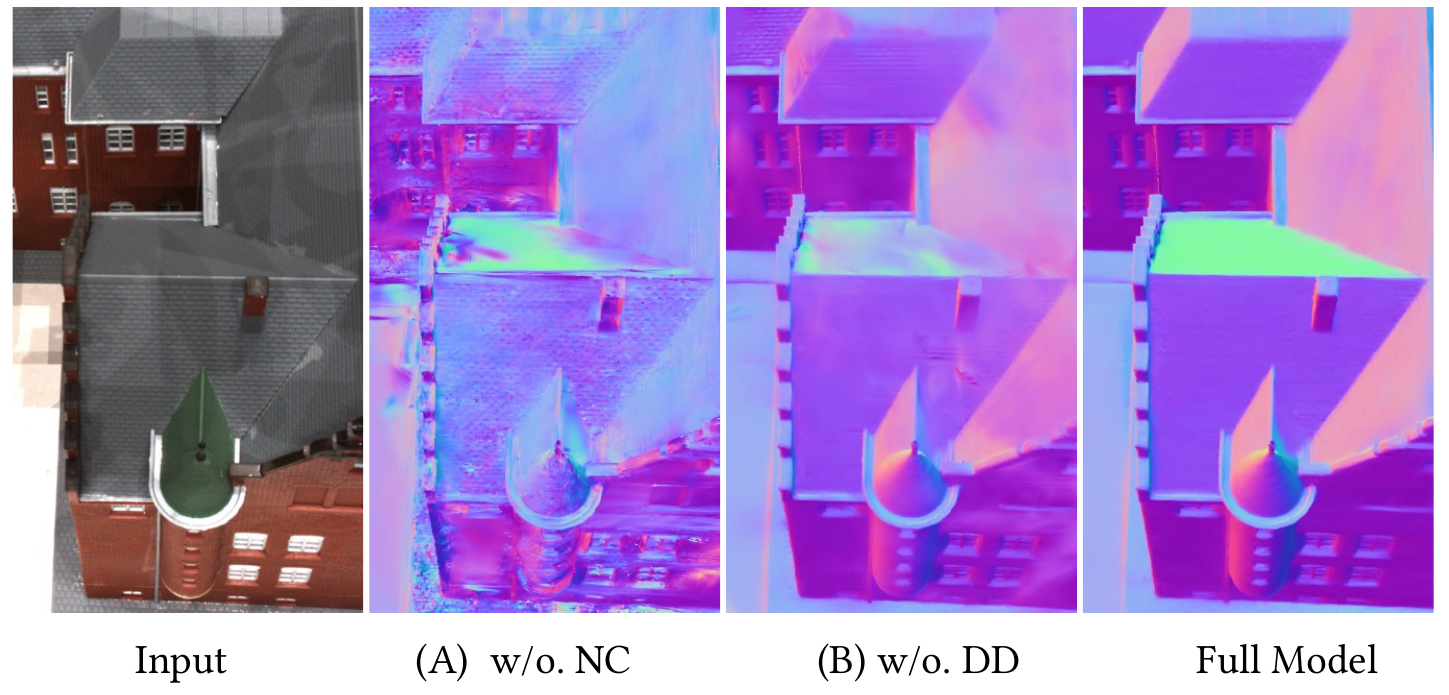}
    \caption{Qualitative studies for the regularization effects. From left to right – input image, surface normals without normal consistency, without depth distortion, and our full model. 
    Disabling the normal consistency loss leads to noisy surface orientations; conversely, omitting depth distortion regularization results in blurred surface normals. The complete model, employing both regularizations, successfully captures sharp and flat features.}
    \label{fig:ablation_reg}
\end{figure}

\subsection{Ablations}
In this section, we isolate the design choices and measure their effect on reconstruction quality, including regularization terms and mesh extraction. 
We conduct experiments on the DTU dataset~\cite{jensen2014large} with 15$k$ iterations and report the reconstruction accuracy, completeness and average reconstruction quality.
The quantitative effect of the choices is reported in ~\tabref{tab:ablation}. {Additional baseline comparisons can be found in Appendix C of the supplemental material.}

\paragraph{Regularization} We first examine the effects of the proposed normal consistency and depth distortion regularization terms. Our model (\tabref{tab:ablation} E) provides the best performance when applying both regularization terms. We observe that disabling the normal consistency (\tabref{tab:ablation} A) can lead to incorrect orientation, as shown in~\figref{fig:ablation_reg} A. Additionally, the absence of depth distortion (\tabref{tab:ablation} B) results in a noisy surface, as shown in~\figref{fig:ablation_reg} B.

\paragraph{Mesh Extraction} We now analyze our choice for mesh extraction. Our full model (\tabref{tab:ablation} E) utilizes TSDF fusion for mesh extraction with median depth. One alternative option is to use the expected depth instead of the median depth. However, it yields worse reconstructions as it is more sensitive to outliers, as shown in~\tabref{tab:ablation} C. Further, our approach surpasses screened Poisson surface reconstruction (SPSR)(\tabref{tab:ablation} D)~\cite{kazhdan2013screened} using 2D Gaussians' center and normal as inputs, due to SPSR's inability to incorporate the opacity and the size of 2D Gaussian primitives. 

\section{Conclusion}
We presented 2D Gaussian splatting, a novel approach for geometrically accurate radiance field reconstruction. We utilized 2D Gaussian primitives for 3D scene representation, facilitating accurate and view consistent geometry modeling and rendering. We proposed two regularization techniques to further enhance the reconstructed geometry. Extensive experiments on several challenging datasets verify the effectiveness and efficiency of our method.

\paragraph{Limitations} 
While our method successfully delivers accurate appearance and geometry reconstruction for a wide range of objects and scenes, 
we also discuss its limitations: First, we assume surfaces with full opacity and extract meshes from multi-view depth maps. This can pose challenges in accurately handling semi-transparent surfaces, such as glass, due to their complex light transmission properties, {as shown in Figure~\ref{fig:supp-1-limitation}}. Secondly, our current densification strategy favors texture-rich over geometry-rich areas, occasionally leading to less accurate representations of fine geometric structures. A more effective densification strategy could mitigate this issue. {Finally, our regularization often involves a trade-off between image quality and geometry, and can potentially lead to over-smoothing in certain regions.}

\begin{acks}
BH and SG are supported by NSFC  \#62172279, \#61932020, Program of Shanghai Academic Research Leader. ZY, AC and AG are supported by the ERC Starting Grant LEGO-3D (850533) and DFG EXC number 2064/1 - project number 390727645.
\end{acks}

\bibliographystyle{ACM-Reference-Format}
\bibliography{main}

\appendix

\section{Details of Depth Distortion}
While Barron~\etal~\cite{barron2022mipnerf360} calculates the distortion loss with samples on the ray, we operate Gaussian primitives, where the intersected depth may not be ordered. To this end, we adopt an $\mathcal{L}_2$ loss and transform the intersected depth $z$ to NDC space to down-weight distant Gaussian primitives, $m = \text{NDC}(z)$, with near and far plane empirically set to 0.2 and 1000. We implemented our depth distortion loss based on \cite{SunSC22_2}, also powered by tile-based rendering. Here we show that the nested algorithm can be implemented in a single forward pass: 
\begin{equation}
\begin{split}
\mathcal{L} &= \sum_{i=0}^{N-1} \sum_{j=0}^{i-1} \omega_i \omega_j (m_i - m_j)^2 \\
&= \sum_{i=0}^{N-1} \omega_i \left(m_i^2 \sum_{j=0}^{i-1} \omega_j  + \sum_{j=0}^{i-1}  \omega_j m_j^2 - 2m_i \sum_{j=0}^{i-1} \omega_j m_j\right)\\
&= \sum_{i=0}^{N-1} \omega_i \left(m_i^2 A_{i-1}  + D^2_{i-1} - 2m_i D_{i-1}\right), 
\end{split}
\end{equation}
where $A_i=\sum_{j=0}^{i}\omega_j$, $D_i = \sum_{j=0}^{i} \omega_jm_j$ and $D^2_i = \sum_{j=0}^{i} \omega_jm_j^2$. 

Specifically, we let $e_i = \left(m_i^2 A_{i-1}+D^2_{i-1}-2m_iD_{i-1}\right)$ so that the distortion loss can be ``rendered'' as $\mathcal{L}_i = \sum_{j=0}^{i}\omega_je_j$. Here, $\mathcal{L}_i$ measures the depth distortion up to the $i$-th Gaussian.  During marching Gaussian front-to-back, we simultaneously accumulate $A_i$, $D_i$ and $D^2_i$, preparing for the next distortion computation $\mathcal{L}_{i+1}$. Similarly, the gradient of the depth distortion can be back-propagated to the primitives back-to-front. Different from implicit methods where $m$ are the pre-defined sampled depth and non-differentiable, we additionally back-propagate the gradient through the intersection $m$, encouraging the Gaussians to move tightly together directly.

\begin{figure}[h]
    \centering
\includegraphics[width=0.98\columnwidth]{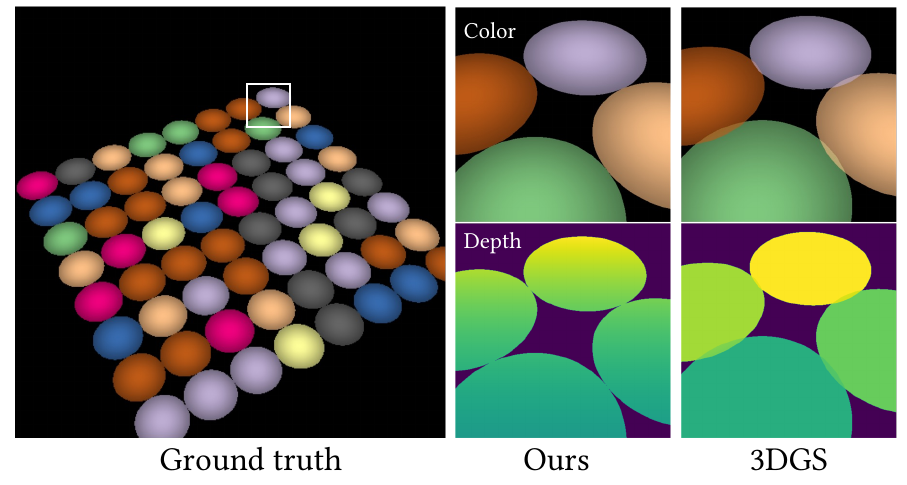}
    \caption{Visualization of a plane tiled by 2D Gaussians. Affine approximation~\cite{zwicker2001surface} adopted in 3DGS~\cite{kerbl3Dgaussians} causes perspective distortion and inaccurate depth, violating normal consistency.}
    \label{fig:perspective}
\vspace{-10pt}
\end{figure}

\section{Depth calculations}
\boldparagraph{Mean depth}There are two optional depth computations used for our meshing process. The mean (expected) depth is calculated by weighting the intersected depth:
\begin{equation}
    \begin{aligned}
    z_\text{mean} = \sum_{i}\omega_iz_i / ({\sum_{i}\omega_i + \epsilon}) \\
\end{aligned}
\label{eq:mean_depth}
\end{equation}
where $\omega_i=T_i\alpha_i\hat{\cG}_i(\bu(\bx)$ is the weight contribution of the $i$-th Gaussian and $T_i=\prod_{j=1}^{i-1} (1 - \alpha_j\,\hat{\cG}_j(\bu(\bx)))$ measures its visibility. It is important to normalize the depth with the accumulated alpha $A=\sum_{i}\omega_i$ to ensure that a 2D Gaussian can be rendered as a planar 2D disk in the depth visualization.

\boldparagraph{Median depth}We compute the median depth as the largest ``visible'' depth, considering $T_i=0.5$ as the pivot for surface and free space:
\begin{equation}
    z_\text{median} = \max\{z_i | T_i > 0.5 \}.
\end{equation}
We find our median depth computation is more robust to \cite{luiten2023dynamic}. When a ray's accumulated alpha does not reach 0.5, while Luiten~\etal sets a default value of 15, our computation selects the last Gaussian, which is more accurate and suitable for training. 

\begin{figure*}[htb]
    \centering
    \resizebox{.99\linewidth}{!}{
    \begin{minipage}{0.325\textwidth}
        \begin{subfigure}[b]{\linewidth}
            \caption{Ground-truth}
            \includegraphics[width=\linewidth]{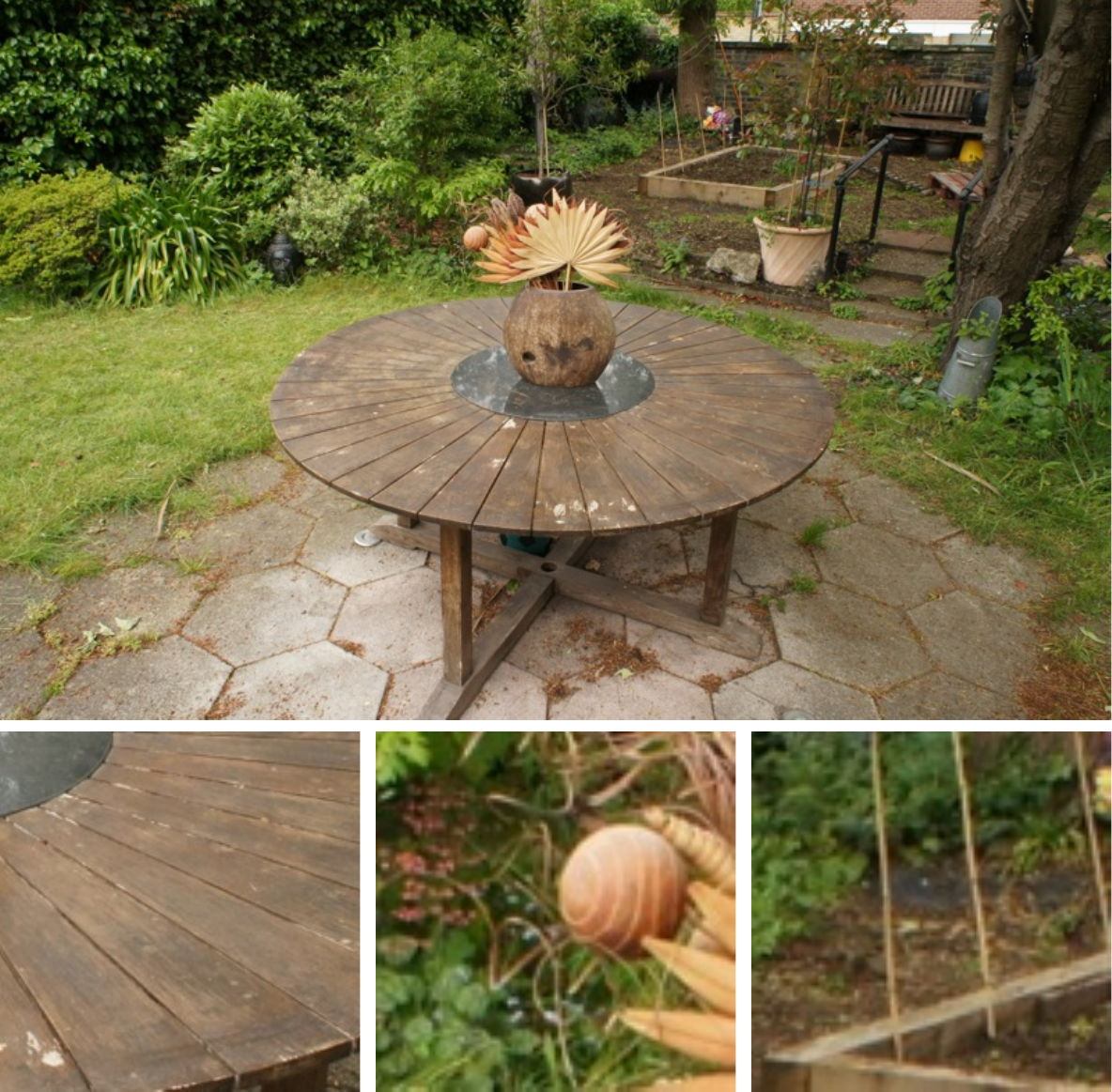}
        \end{subfigure}
        
        \vspace{2pt}
        
        \begin{subfigure}[b]{\linewidth}
            \includegraphics[width=\linewidth]{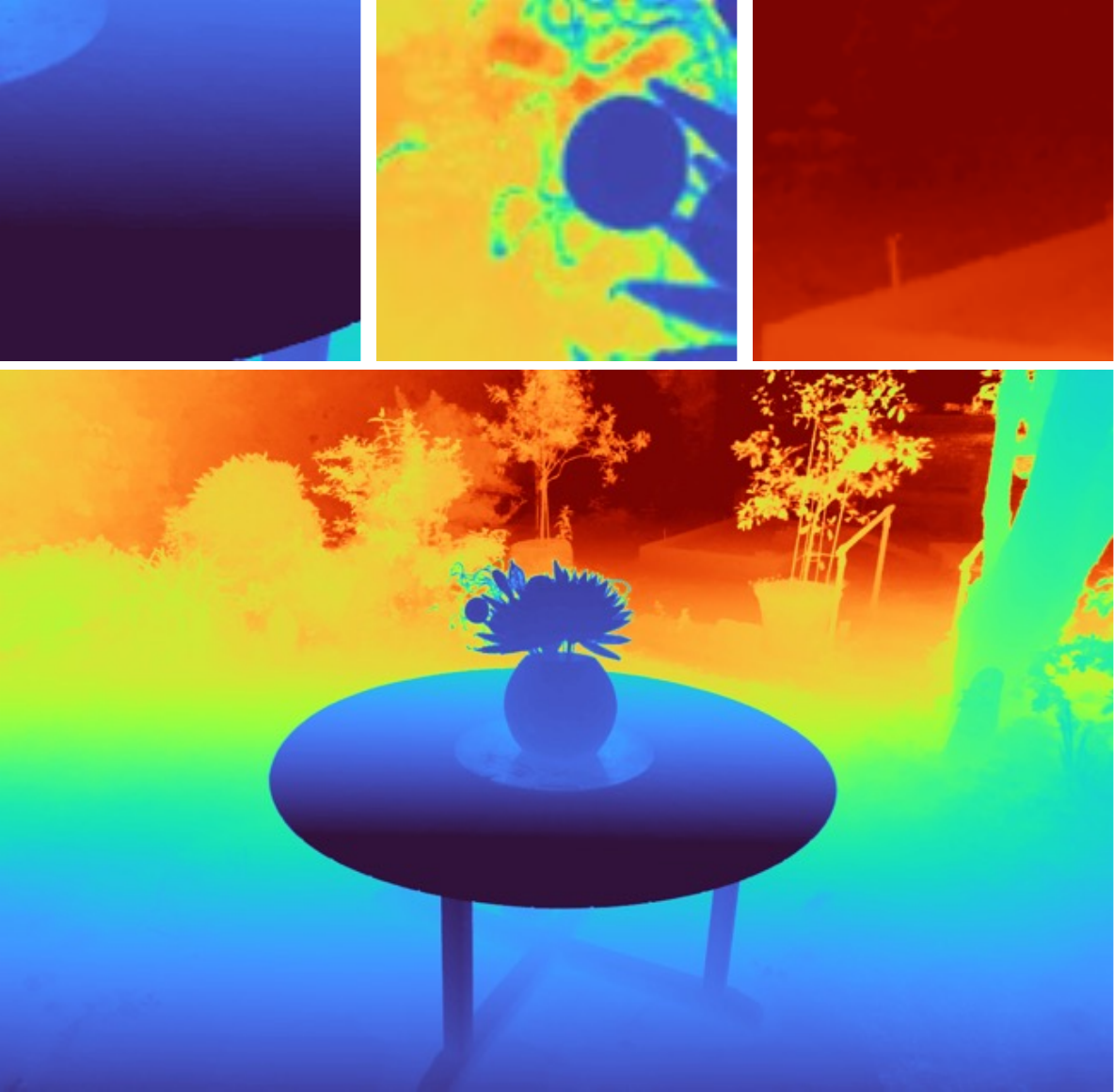}
            \caption{MipNeRF360~\cite{barron2022mipnerf360}, SSIM=0.813}
        \end{subfigure}
    \end{minipage}%
    \hspace{0.0005\textwidth} 
    \vrule width 0.4pt 
    \hspace{0.0005\textwidth} 
    \begin{minipage}{0.325\textwidth}
        \begin{subfigure}[b]{\linewidth}
            \caption{3DGS, normals from depth gradient}
            \includegraphics[width=\linewidth]{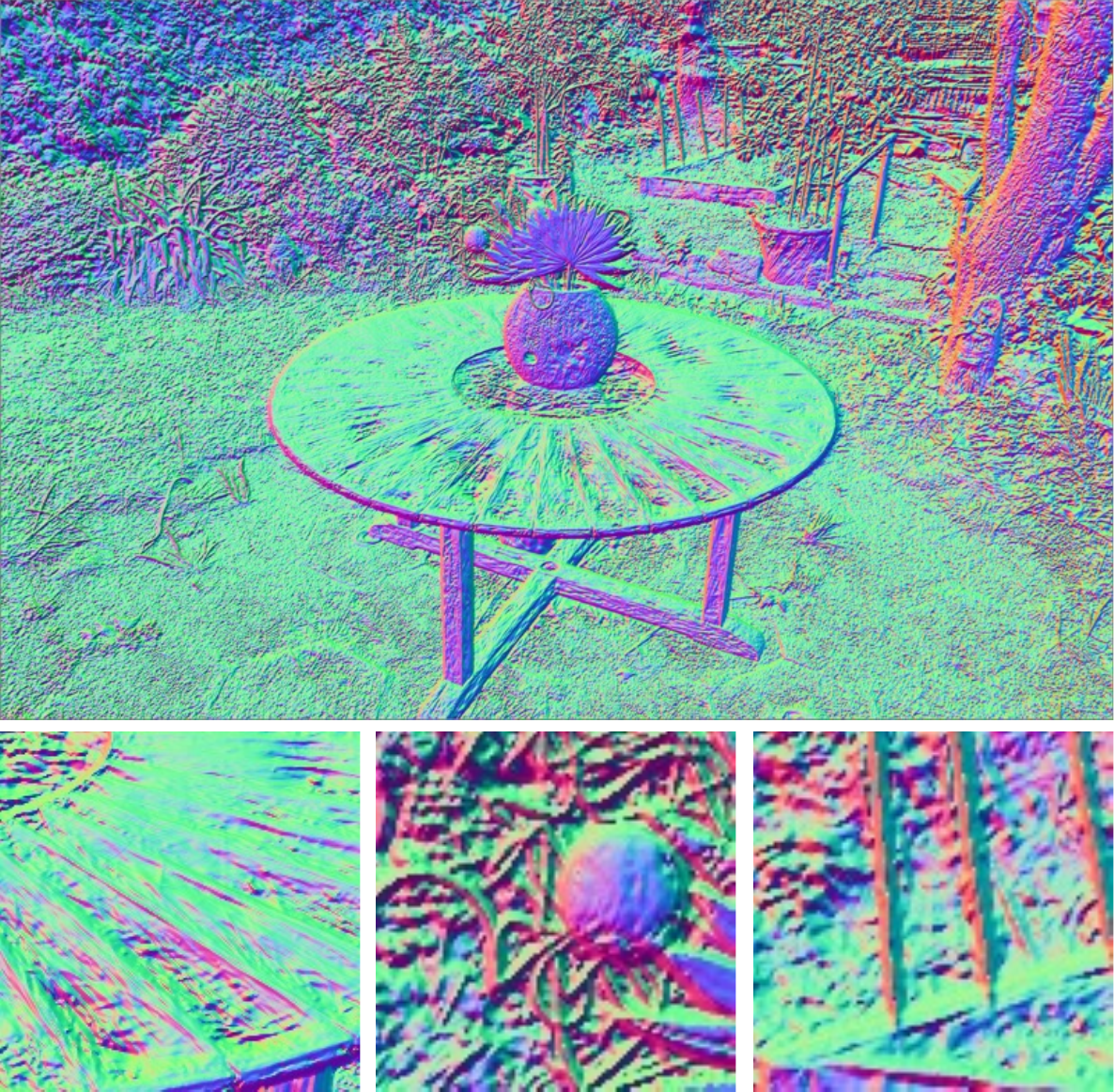}
        \end{subfigure}
        
        \vspace{2pt}
        
        \begin{subfigure}[b]{\linewidth}
            \includegraphics[width=\linewidth]{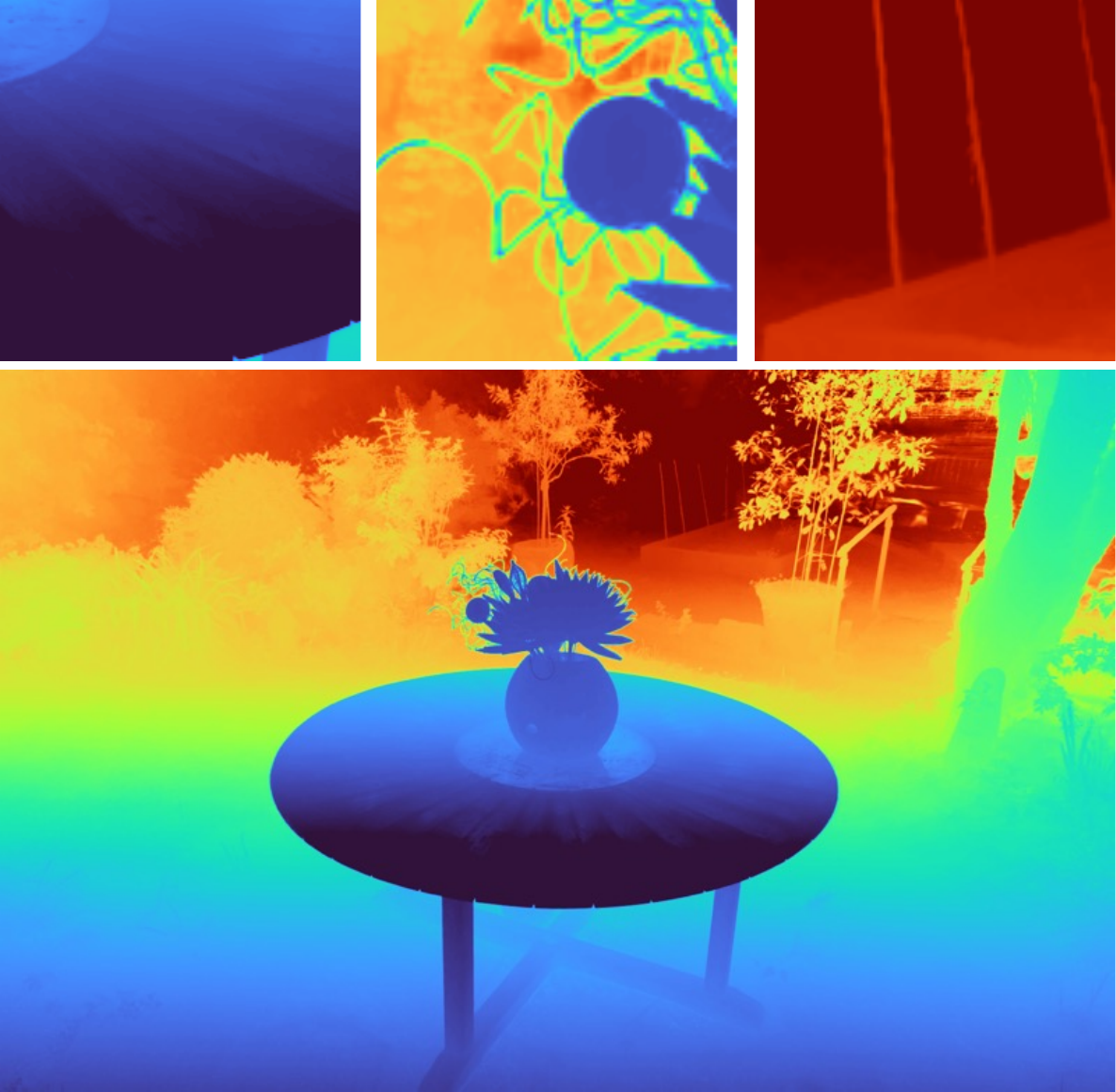}
            \caption{3DGS~\cite{kerbl3Dgaussians}, SSIM=0.834}
        \end{subfigure}
    \end{minipage}%
    \hspace{0.0005\textwidth} 
    \vrule width 0.4pt 
    \hspace{0.0005\textwidth} 
    \begin{minipage}{0.325\textwidth}
        \begin{subfigure}[b]{\linewidth}
            \caption{Our model (2DGS), normals from depth gradient}
            \includegraphics[width=\linewidth]{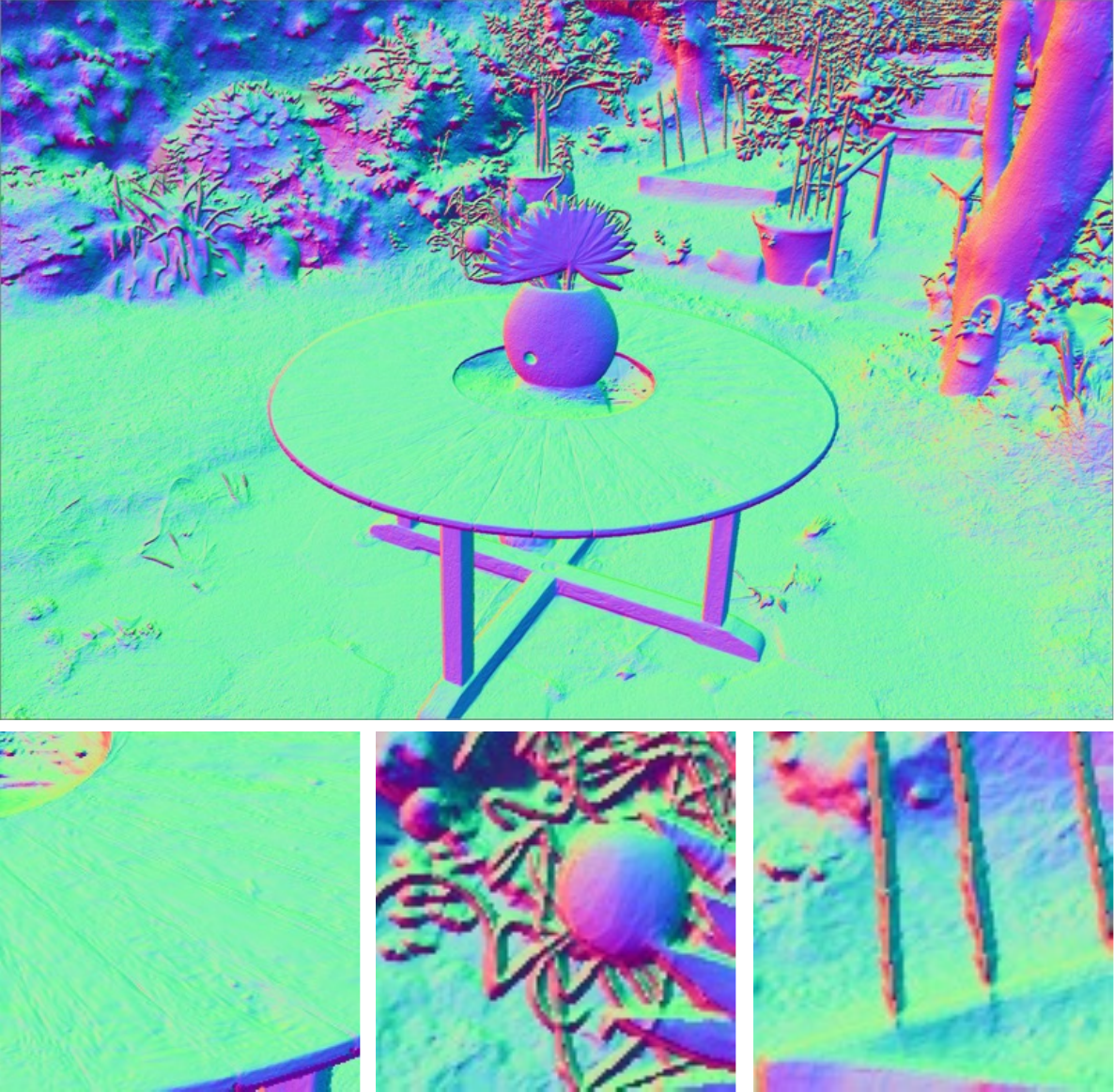}
        \end{subfigure}

        \vspace{2pt}
        
        \begin{subfigure}[b]{\linewidth}
            \includegraphics[width=\linewidth]{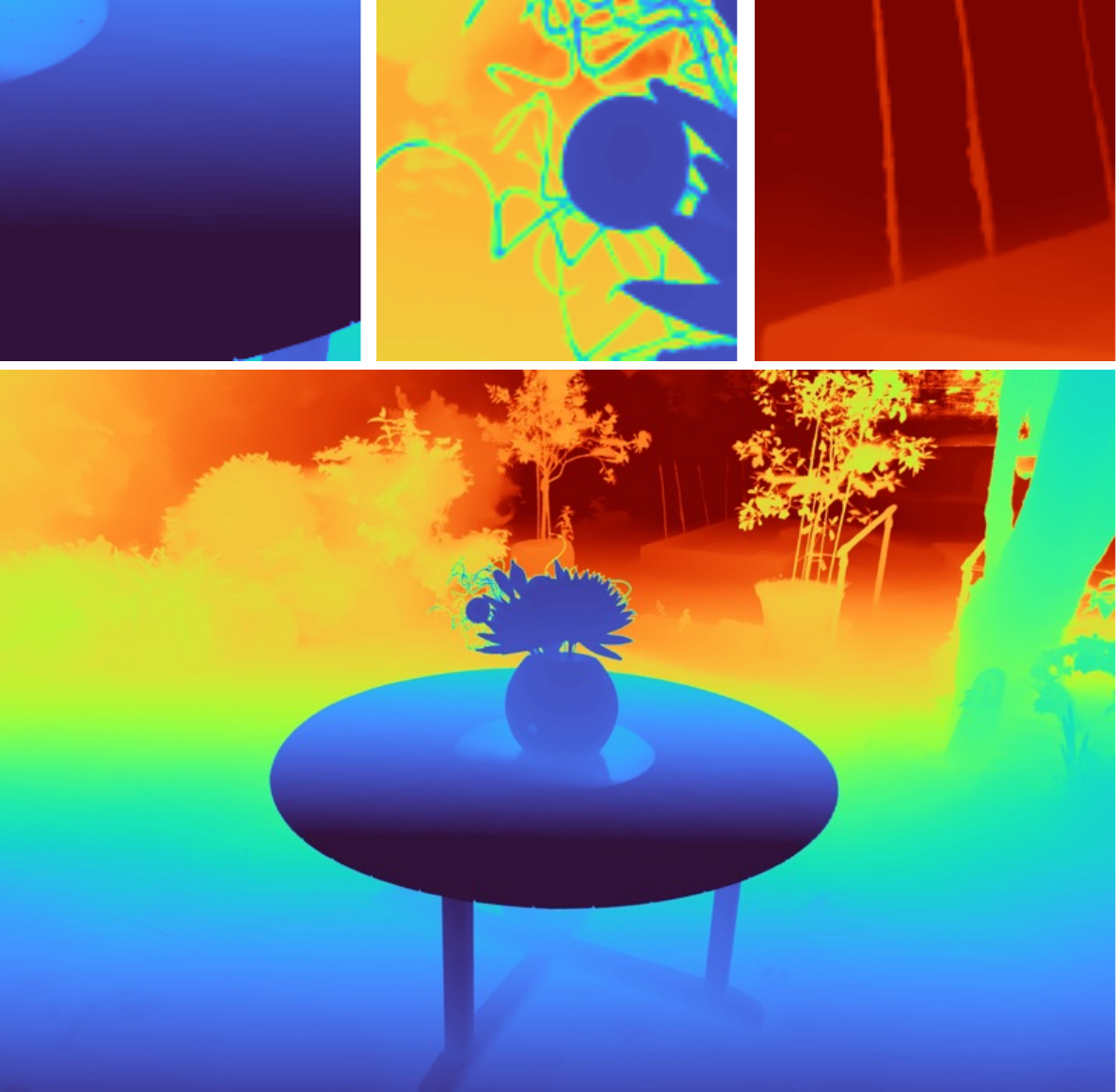}
            \caption{Our model (2DGS), SSIM=0.845}
        \end{subfigure}
    \end{minipage}
    }
    \caption{We visualize the depth maps generated by MipNeRF360~\cite{barron2022mipnerf360}, 3DGS~\cite{kerbl3Dgaussians}, and our method. The depth maps for 3DGS (d) and 2DGS (f) are rendered using Eq.~\ref{eq:mean_depth} and visualized following MipNeRF360. To highlight the surface smoothness, we further visualize the normal estimated from depth gradient using Eq.~\ref{eq:normal_depth} for both 3DGS (c) and ours (e). While MipNeRF360 is capable of producing plausibly smooth depth maps, its sampling process may result in the loss of detailed structures. Both 3DGS and 2DGS excel at modeling thin structures; however, as illustrated in (c) and (e), the depth map of 3DGS exhibits significant noise. In contrast, our approach generates sampled depth points with normals consistent with the rendered normal map (refer to Figure~\ref{fig:teaser}b), thereby enhancing depth fusion during the meshing process.
    }
    \label{fig:supp-depth-normal}
\end{figure*}

\section{Additional baselines}
{In this section, we present additional baselines to ablate the impact of our design choices, as summarized in Table~\ref{tab:supp-additonal-baseline}. Furthermore, we integrate our meshing approach into the comparison against these baselines for a comprehensive analysis. {SuGaR extracts a mesh} from depth points utilizing SPSR (Screen Poisson Surface Reconstruction) during the coarse stage, followed by refinement using a mesh renderer. To assess the effect of this meshing strategy, we substituted SPSR with TSDF using the depth maps, followed by an identical refinement stage. However, we found that the depth map generated from their flat Gaussian intersection is sparse and discontinuous. As a result, the adaptation of TSDF with {its discontinuous depth map yields inferior results.} For 3DGS, we leverage SPSR for mesh generation. {Because the 3D Gaussian lacks a surface normal}, we treat its normal as a trainable parameter~\cite{R3DG2023, liang2023gs} distilled from the depth map, employing the normal consistency regularization. We then utilize all center points for SPSR, resulting in improved overall completion metrics compared to results obtained using TSDF. 

Finally, we conduct ablation experiments on our 2DGS. Notably, 2DGS demonstrates enhanced performance by iteratively integrating components such as TSDF, perspective-correct rasterization, and median depth. For the affine approximation baseline, we utilize 3DGS's rasterization method by configuring one scale of the 3D Gaussian to $1e^{-6}$. While affine approximation already yields promising results, integrating the proposed ray-splat intersection scheme results in more accurate depth map generation under perspective projection, as depicted in Figure~\ref{fig:perspective}, thus enhancing depth fusion performance.
}

\begin{table}[tb]
\caption{Additional baselines on DTU dataset. All the models are trained with 30$k$ iterations.}
\vspace*{-10pt}
\resizebox{0.98\linewidth}{!}{
\begin{tabular}{@{}l|ccc}
 & Accuracy~$\downarrow$ & Completion~$\downarrow$ & Average~$\downarrow$ \\
 \hline
SuGaR  & 1.48 & 1.17 & 1.33 \\
SuGaR + TSDF & 2.47 & 1.90 & 2.18 \\
\hline
3DGS + SPSR (center) & 2.05 & 1.25 & 1.65 \\
3DGS + TSDF (mean) & 1.93 & 1.99 & 1.96 \\
\hline
2DGS + SPSR (center) & 1.25 & \sbest 0.89 & \tbest 1.07 \\
2DGS (affine approx) + TSDF (mean) & \tbest 0.96 & 1.20 &  1.08 \\
2DGS (our rasterizer) + TSDF (mean) & \sbest 0.79 & \tbest 0.98 & \sbest 0.88 \\
2DGS (our rasterizer) + TSDF (median) & \best 0.78 & \best 0.83 & \best 0.80 \\
\end{tabular}
}
\label{tab:supp-additonal-baseline}
\end{table}

\section{Additional Results} 
{
Our 2D Gaussian Splatting method achieves comparable performance even without the need for regularizations, as Table~\ref{tab:nerf} shows.
We have included a detailed breakdown of per-scene metrics for the MipNeRF360 dataset \cite{barron2022mipnerf360} in Table~\ref{tab:mean-mipnerf360-scores}. Additionally, we have provided a comparison of our rendered depth maps with those from 3DGS and MipNeRF360 in Figure~\ref{fig:supp-depth-normal}. 
}

\begin{table}[htb]
\caption{PSNR scores for Synthetic NeRF dataset. Our model achieve comparable performance without using regularizations.}
\centering
\label{tab:nerf_synthetic}
\resizebox{0.95\linewidth}{!}{
\begin{tabular}{l|cccccccc|c}
 & \textbf{Mic} & \textbf{Chair} & \textbf{Ship} & \textbf{Materials} & \textbf{Lego} & \textbf{Drums} & \textbf{Ficus} & \textbf{Hotdog} & \textbf{Mean} \\
\hline
Plenoxels   & 33.26 & 33.98 & 29.62 & 29.14 & 34.10 & 25.35 & 31.83 & 36.81 & 31.76 \\
INGP-Base   & 36.22 & 35.00 & 31.10 & 29.78 & 36.39 & 26.02 & 33.51 & 37.40 & 33.18 \\
Mip-NeRF    & 36.51 & 35.14 & 30.41 & 30.71 & 35.70 & 25.48 & 33.29 & 37.48 & 33.09 \\
3DGS        & 35.36 & 35.83 & 30.80 & 30.00 & 35.78 & 26.15 & 34.87 & 37.72 & 33.32 \\
Ours        & 35.09 & 35.05 & 30.60 & 29.74 & 35.10 & 26.05 & 35.57 & 37.36 & 33.07 \\
\end{tabular}
}
\label{tab:nerf}
\end{table}

\begin{table}[htb]
\caption{PSNR scores for TnT dataset.}
\vspace*{-10pt}
\resizebox{0.98\linewidth}{!}{
\begin{tabular}{l|cccccc|c}
 & Barn & Caterpillar & Courthouse & Ignatius & Meetingroom & Truck & Mean\\
\hline
SuGaR & 28.63 &	23.27 &	23.33 &	20.72 &	25.47 &	24.40 &	24.16 \\
3DGS & 27.99 &	24.82 &	23.33 &	23.95 &	26.89 &	25.01 &	25.33 \\
Ours & 28.79 &	24.23 &	23.51 &	23.82 &	26.15 &	26.85 & 25.56 \\
\end{tabular}
}
\label{tab:mean-tnt-scores}
\end{table}

\begin{table}[htb]
\caption{PSNR$\uparrow$, SSIM$\uparrow$, LIPPS$\downarrow$ scores for MipNeRF360 dataset.}
\resizebox{\linewidth}{!}{
\begin{tabular}{l|ccccc|cccc|c}
 & bicycle & flowers & garden & stump & treehill & room & counter & kitchen & bonsai & mean\\
\hline
SugaR & 23.34 & 19.54 & 25.40 & 25.07 & 21.30 & 29.97 & 27.56 & 29.41 & 30.77 & 25.82\\
3DGS & 25.24 & 21.52 & 27.41 & 26.55 & 22.49 & 30.63 & 28.70 & 30.32 &  31.98 & 27.20 \\
Ours & 24.87 & 21.15 & 26.95 & 26.47 & 22.27 & 31.06 & 28.55 & 30.50 & 31.52 & 27.03 \\
\hline
SuGaR & 0.634 & 0.499 & 0.762 & 0.705 & 0.546 & 0.904 & 0.885 & 0.902 & 0.933 & 0.752 \\
3DGS & 0.771 & 0.605 & 0.868 & 0.775 & 0.638 &0.914 & 0.905 & 0.922 & 0.938 & 0.815 \\
Ours & 0.752 & 0.588 & 0.852 & 0.765 & 0.627 & 0.912 & 0.900 & 0.919 & 0.933 & 0.805 \\
\hline
SuGaR & 0.354 & 0.407 & 0.240 & 0.325 & 0.452 & 0.259 & 0.244 & 0.178 & 0.220 & 0.298 \\
3DGS & 0.205& 0.336& 0.103& 0.210& 0.317& 0.220& 0.204& 0.129& 0.205& 0.214 \\
Ours & 0.218 & 0.346 & 0.115 & 0.222 & 0.329 & 0.223 & 0.208 & 0.133 & 0.214 & 0.223 \\
\end{tabular}
}
\label{tab:mean-mipnerf360-scores}
\end{table}
\begin{figure*}[t]
    \centering
    \includegraphics[width=0.95\linewidth]{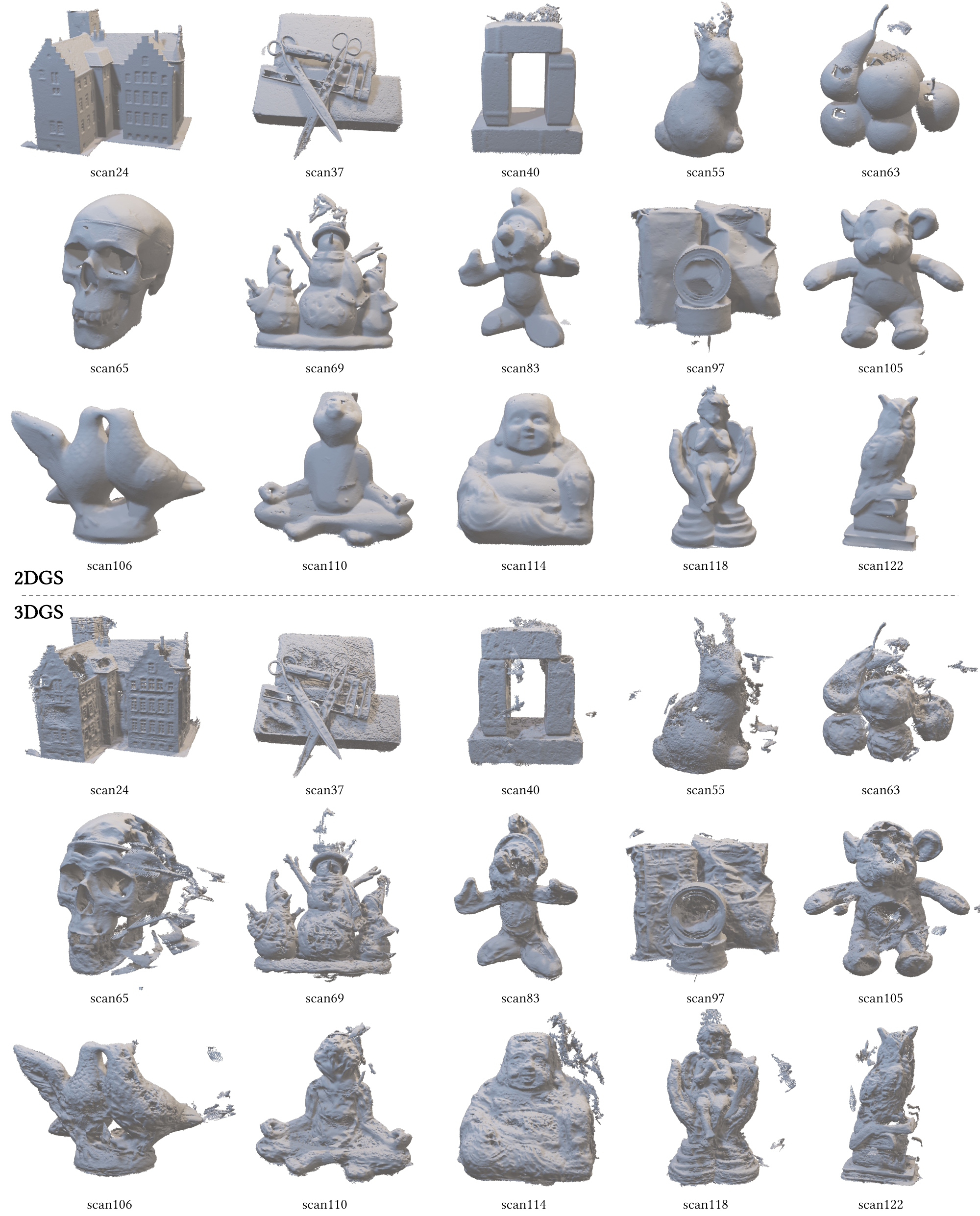}
    \caption{Comparison of surface reconstruction using our 2DGS and 3DGS~\cite{kerbl3Dgaussians}. Meshes are extracted by applying TSDF to the depth maps.}
    \label{fig:supp-DTU}
\end{figure*}

\begin{figure*}[t]
    \centering
    \includegraphics[width=0.97\linewidth]{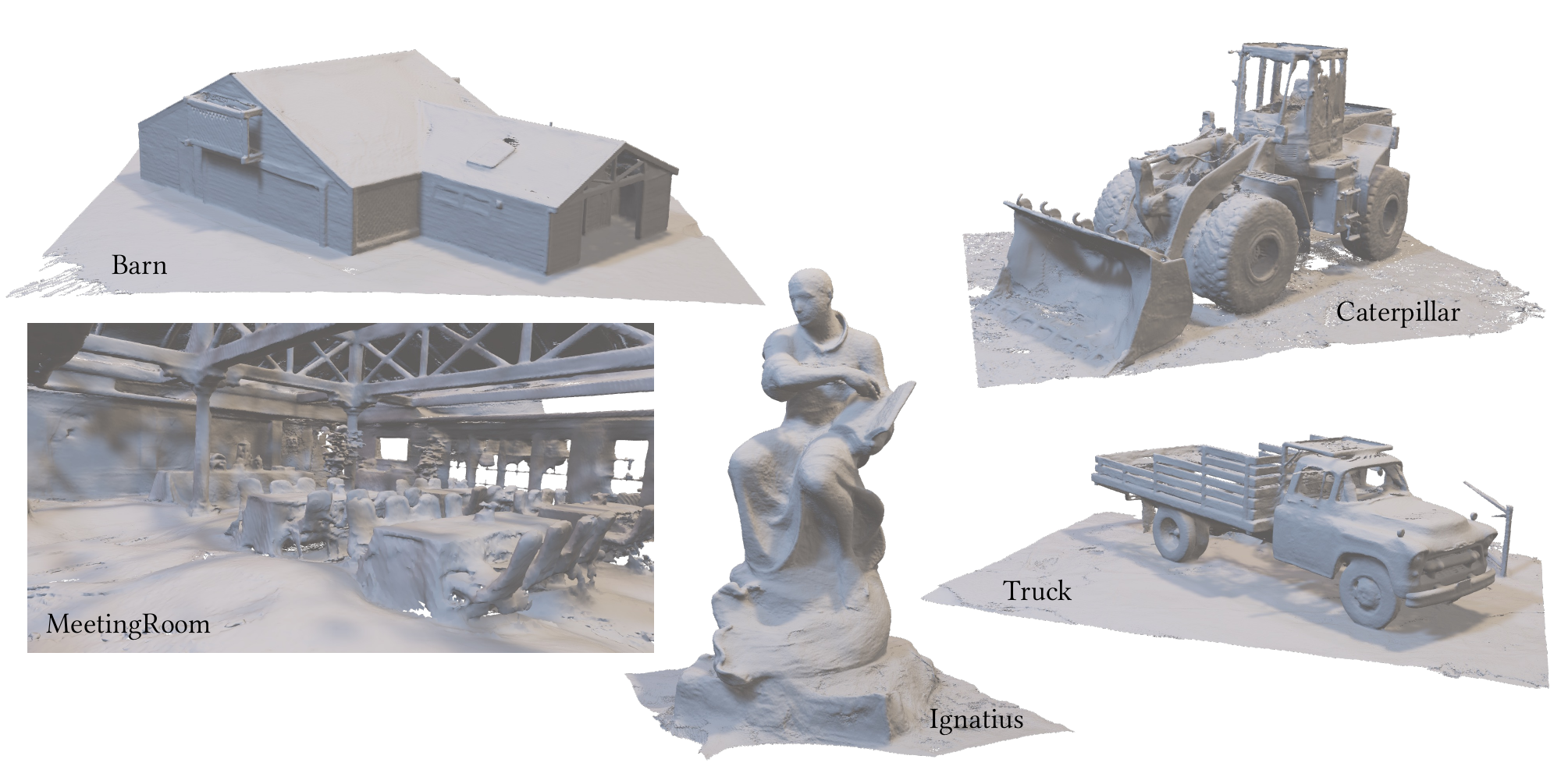}
    \vspace{-0.5cm}
    \caption{Qualitative studies for the Tanks and Temples dataset~\cite{Knapitsch2017}.}
    \label{fig:supp-TnT}
\end{figure*}

\begin{figure*}[t]
    \centering
    \includegraphics[width=0.97\linewidth]{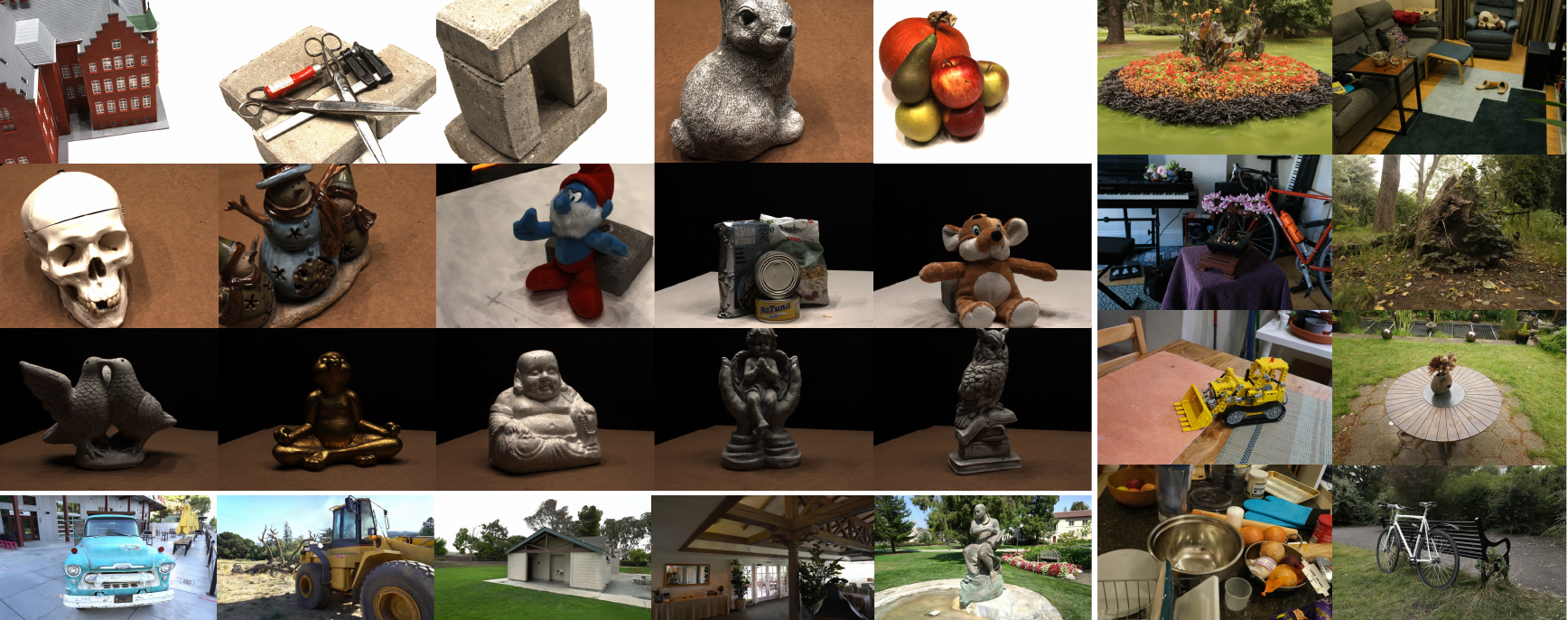}
    \vspace{-0.2cm}
    \caption{Appearance rendering results from reconstructed 2D Gaussian disks, including DTU, TnT, and Mip-NeRF360 datasets.}
    \label{fig:supp-rendering}
\end{figure*}

\begin{figure*}[t]
    \centering
    \includegraphics[width=0.97\linewidth]{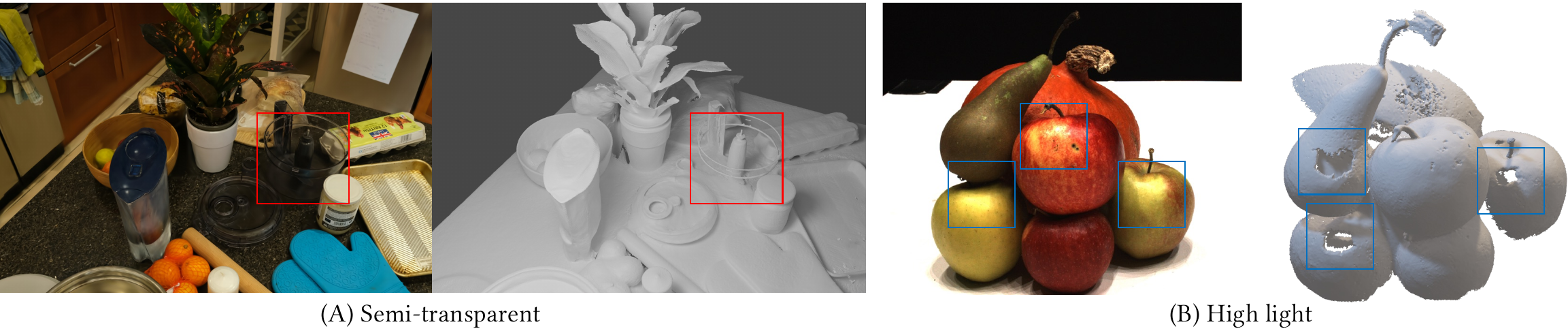}
    \vspace{-0.2cm}
    \caption{Illustration of limitations: Our 2DGS struggles with the accurate reconstruction of semi-transparent surfaces, for example, the glass shown in example (A). Moreover, our method tends to create holes in areas with high light intensity, as shown in (B).}
    \label{fig:supp-1-limitation}
\end{figure*}
\end{document}